\documentclass{article}

\usepackage[square,numbers]{natbib}

\usepackage[preprint]{neurips_2025}

\usepackage[utf8]{inputenc} 
\usepackage[T1]{fontenc}    
\usepackage{hyperref}       
\usepackage{url}            
\usepackage{booktabs}       
\usepackage{amsfonts}       
\usepackage{nicefrac}       
\usepackage{microtype}      
\usepackage{graphicx}       
\usepackage{algorithm}      
\usepackage{algorithmic}    
\usepackage{amsmath}        
\usepackage{multirow}       
\usepackage{subcaption}     
\usepackage[table]{xcolor}
\newcommand{\phy}[1]{\textcolor{black}{#1}}

\usepackage{listings}
\definecolor{codegreen}{rgb}{0,0.6,0}
\definecolor{codegray}{rgb}{0.5,0.5,0.5}
\definecolor{codepurple}{rgb}{0.58,0,0.82}
\definecolor{backcolour}{rgb}{0.95,0.95,0.92}

\lstdefinestyle{mystyle}{
    backgroundcolor=\color{backcolour},   
    commentstyle=\color{codegreen},
    keywordstyle=\color{magenta},
    numberstyle=\tiny\color{codegray},
    stringstyle=\color{codepurple},
    basicstyle=\ttfamily\footnotesize,
    breakatwhitespace=false,         
    breaklines=true,                 
    captionpos=b,                    
    keepspaces=true,                 
    numbers=left,                    
    numbersep=5pt,                  
    showspaces=false,                
    showstringspaces=false,
    showtabs=false,                  
    tabsize=2
}

\definecolor{remark}{rgb}{1,.5,0} 
\definecolor{citecolor}{rgb}{0,0.443,0.737} 
\definecolor{linkcolor}{RGB}{219, 48, 122} 
\definecolor{cyan}{rgb}{0.831,0.901,0.945}
\definecolor{myblue}{RGB}{79,113,190}

\definecolor{best_color}{rgb}{1, 0.7, 0.7}
\definecolor{second_color}{rgb}{1, 0.85, 0.7}
\definecolor{third_color}{rgb}{1, 1, 0.7}

\newcommand{\best}[1]{\cellcolor{best_color}{#1}}
\newcommand{\second}[1]{\cellcolor{second_color}{#1}}
\newcommand{\third}[1]{\cellcolor{third_color}{#1}}

\hypersetup{
    colorlinks=true,%
    citecolor=citecolor,%
    filecolor=citecolor,%
    linkcolor=linkcolor,%
    urlcolor=magenta
}

\title{
EX-4D: EXtreme Viewpoint 4D Video Synthesis via Depth Watertight Mesh}

\author{
Tao Hu\thanks{Equal contribution}~ \thanks{Corresponding author} \qquad
Haoyang Peng\footnotemark[1] \qquad
Xiao Liu \qquad
Yuewen Ma
\\\\
Pico, Bytedance \\\\
\url{https://tau-yihouxiang.github.io/projects/EX-4D/EX-4D.html}
}

\begin{document}

\maketitle

\begin{figure}[ht]
    \centering
    \includegraphics[width=\linewidth]{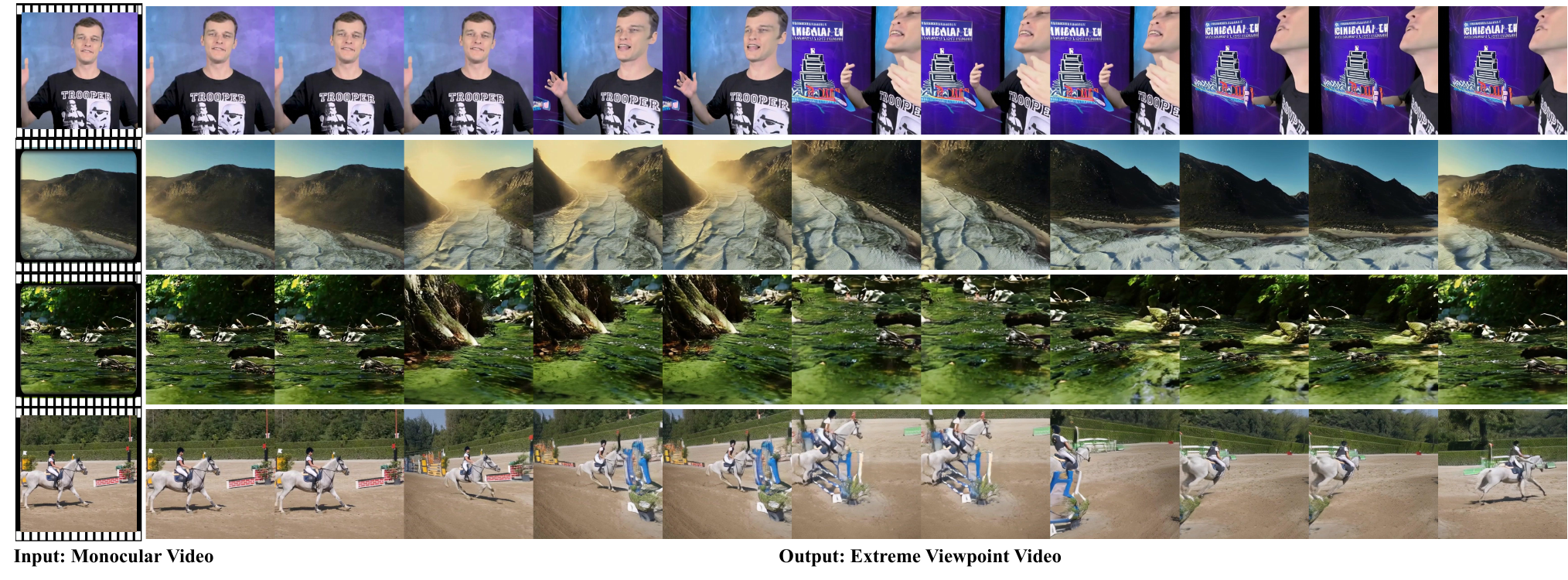}
    \caption{Our EX-4D framework takes a monocular video as input and generates high-quality 4D videos under extreme viewpoint. By leveraging the proposed Depth Watertight Mesh representation, it effectively handles occlusions in boundaries and ensures geometric consistency, enabling visually coherent and realistic results.}
    \label{fig:teaser}
\end{figure}

\begin{abstract}
Generating high-quality camera-controllable videos from monocular input is a challenging task, particularly under extreme viewpoint. Existing methods often struggle with geometric inconsistencies and occlusion artifacts in boundaries, leading to degraded visual quality. In this paper, we introduce EX-4D, a novel framework that addresses these challenges through a Depth Watertight Mesh representation. The representation serves as a robust geometric prior by explicitly modeling both visible and occluded regions, ensuring geometric consistency in extreme camera pose. To overcome the lack of paired multi-view datasets, we propose a simulated masking strategy that generates effective training data only from monocular videos. Additionally, a lightweight LoRA-based video diffusion adapter is employed to synthesize high-quality, physically consistent, and temporally coherent videos. Extensive experiments demonstrate that EX-4D outperforms state-of-the-art methods in terms of physical consistency and extreme-view quality, enabling practical 4D video generation.
\end{abstract}

\section{Introduction}
Recent advancements in video generative models \cite{SVD,CogVideoX,HunyuanVideo,Step-Video-T2V,Wan} have enabled high-quality, controllable video synthesis from diverse prompts including text, images, and videos. Within this rapidly evolving field, camera-controllable (or called 4D) video generation \cite{CameraCtrl,GenXD,ReconX} has emerged as a particularly significant direction, allowing viewers to experience dynamic scenes from multiple viewpoints by simultaneously modeling spatial, temporal, and viewpoint dimensions. This capability represents a fundamental breakthrough for applications in mixed reality experiences, free-viewpoint video systems, and immersive 3D movies production.

However, generating camera-controllable videos for dynamic scenes with extreme viewpoints (e.g., ranging from $-90$° to $90$°) remains challenging. Current approaches fall into two categories:
1) Camera-based guidance \cite{CameraCtrl,Syncammaster,Ac3d,ReCamMaster,Zero123} use camera parameters as implicit conditions, encoding position and orientation through ray maps \cite{Plucker}, positional embeddings \cite{NeRF}, or relative pose prompts. While these can generate videos with varying camera poses, they lack physical-consistent controllability and require extensive multi-view datasets with accurate camera calibration.
2) Geometry-based guidance \cite{DaS,TrajectoryCrafter,TrajectoryAttention} leverage explicit 3D representations to enable viewpoint control. These approaches first reconstruct 3D geometry (e.g., point clouds, meshes) from input frames using techniques like MVSNet \cite{MVSNet} or recent Pointmap \cite{DUSt3R,Monst3R,Cut3R,VGGT}, then render these representations from target camera parameters to guide video generation. While they reduce dependence on camera-calibrated multi-view training data \cite{TrajectoryCrafter}, these methods face fundamental limitations in representing occluded regions. Their incomplete geometry modeling causes significant artifacts in the boundaries when synthesizing extreme viewpoints, particularly in areas that transition from hidden to visible as the camera moves.

To address these limitations, we propose \textbf{EX-4D}, a novel framework for transforming monocular videos into \textbf{EX}treme viewpoint \textbf{4D} videos (see Fig.~\ref{fig:teaser}). Our approach bridges the gap between camera-based and geometry-based methods, addressing the core challenge of synthesizing convincing extreme-view 4D videos from monocular input without requiring camera-calibrated multi-view training data while enabling physically consistent viewpoint control, which implies boundary occlusion continuity and temporal appearance coherence.

The core of our framework is the Depth Watertight Mesh (DW-Mesh) representation, which serves as a robust geometric prior to guide the video generation process. Unlike traditional surface reconstruction, which struggles with sparse visibility across viewpoints, the DW-Mesh explicitly models both visible surfaces and occluded boundaries, ensuring geometric consistency even under extreme camera movements. This representation provides reliable and complete masks for every viewpoint, effectively handling occlusions in boundaries through its watertight structure.

To address the lack of multi-view training data, we propose a simulated masking strategy that creates effective training samples from monocular videos. This approach uses two complementary techniques: (1) Rendering Mask Generation, which creates visibility masks from our DW-Mesh to simulate novel viewpoint occlusions; and (2) Tracking Mask Generation, which ensures temporal consistency by tracking points across frames. This strategy eliminates the need for expensive multi-view data collection while effectively simulating extreme viewpoint challenges.

Finally, guided by the DW-Mesh priors, a lightweight LoRA-based video diffusion adapter synthesizes high-quality videos with enhanced temporal coherence. This adapter efficiently integrates the geometric information from the DW-Mesh with pre-trained video diffusion models, producing visually coherent and realistic results while keeping computational requirements manageable. Our experiments demonstrate that EX-4D consistently outperforms state-of-the-art methods across different metrics and viewpoint ranges, with the performance gap widening as camera angles become more extreme. User studies further confirm that our method produces more realistic and physically consistent videos, particularly for challenging viewpoints ranging from $-90$° to $90$°.

In summary, our main contributions are:
\begin{enumerate}
    \item We introduce the DW-Mesh representation that models both visible and hidden regions, maintaining geometric consistency for extreme viewpoints.
    \item We develop a simulated masking strategy that enables training without requiring multi-view video datasets.
    \item We design a lightweight LoRA-based adapter with only 1\% trainable parameters that efficiently combines geometric information with pre-trained video models.
    \item Our experiments show EX-4D outperforms existing methods, especially for extreme camera angles and complex occlusions.
\end{enumerate}

\section{Related Work}

\paragraph{3D and 4D Scene Reconstruction.}
Recent advances include neural representations like NeRF~\cite{NeRF}, efficient methods such as 3D Gaussian Splatting~\cite{3DGS}, and dynamic scene modeling with Shape-of-Motion~\cite{SOM}. Approaches like DUSt3R~\cite{DUSt3R}, X-Ray~\cite{X-Ray}, CUT3R~\cite{Cut3R}, and VGGT~\cite{VGGT} have improved efficiency by reconstructing from uncalibrated images. However, these methods often struggle with occlusions and dynamic scenes. Our DW-Mesh explicitly models occluded regions to ensure geometric consistency during extreme viewpoint synthesis.

\paragraph{Video Diffusion Models.}
The field has evolved from early approaches like Make-A-Video~\cite{Make-A-Video} and Gen-1~\cite{Gen-1} to more sophisticated models. SVD~\cite{SVD} and VideoCrafter~\cite{VideoCrafter1,VideoCrafter2} enhanced temporal coherence, while large-scale models such as Hunyuan Video~\cite{HunyuanVideo}, CogVideoX~\cite{CogVideoX}, and Wan 2.1~\cite{Wan} achieve impressive spatiotemporal consistency. Our framework builds on these capabilities to enable extreme-angle video generation with robust geometric and temporal coherence.

\paragraph{Camera and Motion Control.}
Various approaches enable camera movement in video synthesis. CameraCtrl~\cite{CameraCtrl} uses camera parameter encoding but struggles with extreme viewpoints. GCD~\cite{GCD} employs pose embeddings but requires domain-specific training. TrajectoryCrafter~\cite{TrajectoryCrafter} enables camera redirection using point clouds but has reconstruction issues. ReCamMaster~\cite{ReCamMaster} extends T2V models but needs multi-camera training data. Other approaches like MotionCtrl~\cite{Motionctrl}, AnimateDiff~\cite{AnimateDiff,SparseCtrl}, and DragNUWA~\cite{DragNUWA} support basic camera effects without proper geometric understanding. Our approach addresses these limitations through DW-Mesh representation, enabling high-quality novel view synthesis from monocular videos without multi-view training data.

\section{Our Approach}
The goal of our EX-4D framework is to generate novel-view videos $V_t = \{I_t\}_{t=1}^{T}$ from an input monocular video $V_s = \{I_t\}_{t=1}^{S}$ and a target camera trajectory $\{P_t\}_{t=1}^{T}$. It consists of three key steps: (1) constructing a DW-Mesh as a geometric prior to handle occlusions in boundaries, (2) generating training masks to simulate novel view occlusions using monocular videos, and (3) using a lightweight video diffusion adapter to produce physically or geometrically consistent and temporally coherent videos.

\subsection{Depth Watertight Mesh}
Existing 3D representations for novel view synthesis typically focus on visible surfaces while neglecting occluded regions, leading to artifacts when rendering from extreme viewpoints. Our DW-Mesh addresses this fundamental limitation by implementing a geometric structure that maintains both visible and hidden surfaces through a watertight formulation. This technical design choice enables unified handling of scene topology across arbitrary camera positions without requiring explicit multi-view supervision.

\subsubsection{DW-Mesh Construction}
\begin{figure}[t]
    \centering
        \includegraphics[width=0.75\textwidth]{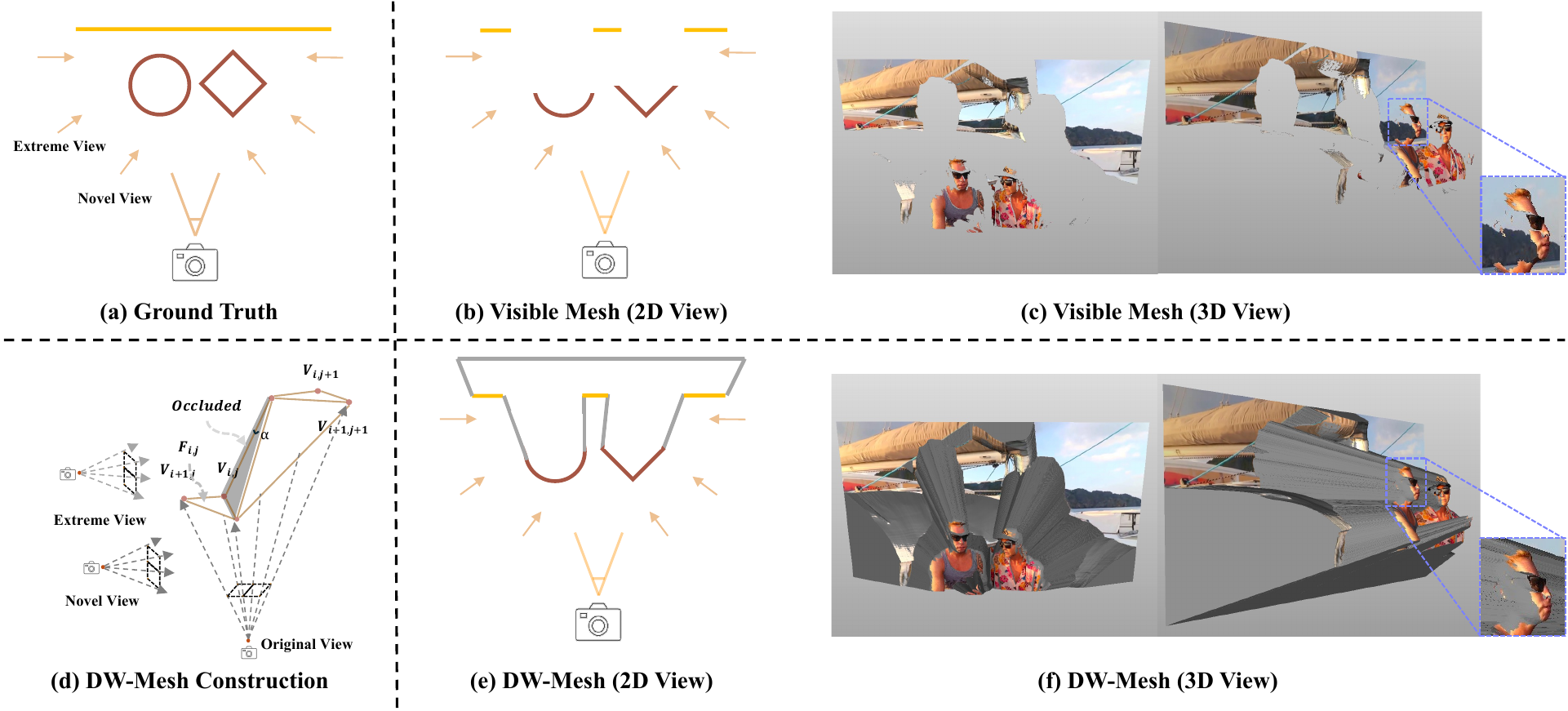}
        \caption{Illustration of DW-Mesh construction. (a) Ground Truth: The original scene with complete geometry. (b) Visible Mesh (2D View): 3D reconstructed visible mesh representation showing only the visible regions. (c) Visible Mesh (3D View): 3D visualization of the visible mesh, highlighting missing occluded regions with wrong physical boundaries. (d) DW-Mesh Construction: Our proposed Depth Watertight Mesh explicitly models both visible and occluded regions, ensuring geometric consistency. (e) DW-Mesh (2D View): 2D representation of the DW-Mesh, showing the inclusion of occluded areas. (f) DW-Mesh (3D View): 3D visualization of the DW-Mesh, demonstrating its watertight structure and ability to handle occlusions in boundaries to maintain physical consistency.}
        \label{fig:dw_mesh_construction}
\end{figure}

As shown in Fig. \ref{fig:dw_mesh_construction}, for each video frame $I_t$, we construct its DW-Mesh $M_t=\{V, F, T, O\}$, where $V$ represents vertices, $F$ denotes faces, $T$ represents mesh textures, and $O$ indicates whether faces are occluded. The construction process involves the following steps:

\paragraph{Vertex.} We compute per-frame depth maps $D_t$ using a pre-trained depth estimation model \cite{DepthCrafter}. Each pixel $(i,j)$ with depth value $D_{i,j}$ is unprojected into 3D space to form a vertex $V_{i,j}$:
\begin{equation}
    V_{i,j} = o + D_{i,j} \cdot r_{i,j},
\end{equation}
where $o$ is the canonical camera origin, and $r_{i,j}$ is the ray direction for pixel $(i,j)$. We adopt canonical camera space to bypass the need for  camera calibration \cite{Splatter_a_video}.  \phy{To ensure DW-Mesh remains a closed surface when unprojected into 3D space under arbitrary viewpoints, we implement a boundary padding strategy by assigning $D_\text{max}$ as the depth value of all frame-border pixels.} 

\paragraph{Face.} A triangular face is constructed by connecting adjacent vertices. For each 2x2 grid of pixels, two triangular faces are formed:
\begin{align}
    F_{i,j,1} &= \{(i,j), (i+1,j), (i,j+1)\}, \\
    F_{i,j,2} &= \{(i+1,j), (i+1,j+1), (i,j+1)\}.
\end{align}
For simplicity, we denotes $F_{i,j}$ as an arbitrary face. \phy{Two additional faces $\{(0,0), (0,W), (H,0)\}$ and $\{(H,0), (H,W), (0,W)\}$ are added to ensure watertight mesh construction.}


\paragraph{Occlusion and Texture.} 
\label{sec:tex and occ}

Rather than directly assigning pixel color as texture, which can lead to artifacts in depth boundary areas, we add an additional occlusion attribute. \phy{For each face, we perform geometric validation through minimum face angle analysis and depth discontinuity detection. Faces with a minimum angle less than $\delta_\text{angle}$ or with large depth discontinuities $\Delta D$ are considered geometrically degenerate and marked as occluded:}

\begin{equation}
    O_{i,j} = 
    \begin{cases}
      1, & \text{if } \text{Min}(\angle(F_{i,j})) < \delta_\text{angle} \text{ or } \Delta D > \delta_{\text{depth}}, \\  
      0, & \text{otherwise}.
    \end{cases}
\end{equation}

\phy{The texture value $T_{i,j}$ is then assigned as:}

\begin{equation}
    T_{i,j} = 
    \begin{cases}
      [0,0,0], & \text{if } O_{i,j} = 1, \\
      C_{i,j}, & \text{otherwise},
    \end{cases}
\end{equation}

where $C_{i,j}\in\mathbb{R}^3$ is the color of \phy{corresponding} pixel $(i,j)$.


This process produces a watertight mesh $M_t=\{V, F, T, O\}$ for each frame, capturing both visible and occluded regions. The watertight property is crucial for handling extreme viewpoint, ensuring geometric consistency even for previously unseen parts of the scene. The resulting DW-Mesh $M = \{M_t\}_{t=1}^T$ provides a comprehensive geometric representation across time, enabling reliable rendering from arbitrary viewpoints.

\subsubsection{DW-Mesh Rendering}
The DW-Mesh is rendered using rasterization without additional lighting effects from the target camera trajectory $\{P_t\}_{t=1}^{T}$ to produce a color video from the mesh face texture $T$ and a mask video from the occlusion attribute $O$. These outputs serve as geometric priors, conditioning the video diffusion module to synthesize novel-view frames with improved visual consistency and geometric accuracy, even under challenging camera movements. Although the watertight mesh may occlude visible and valid surfaces in large or extreme viewpoints after mesh rendering, our subsequent adapter and video diffusion model effectively address this issue, generating reasonable and visually coherent results for the occluded areas.

\subsection{Mask Generation for Training}
\begin{figure}[ht]
    \centering
    \includegraphics[width=0.8\linewidth]{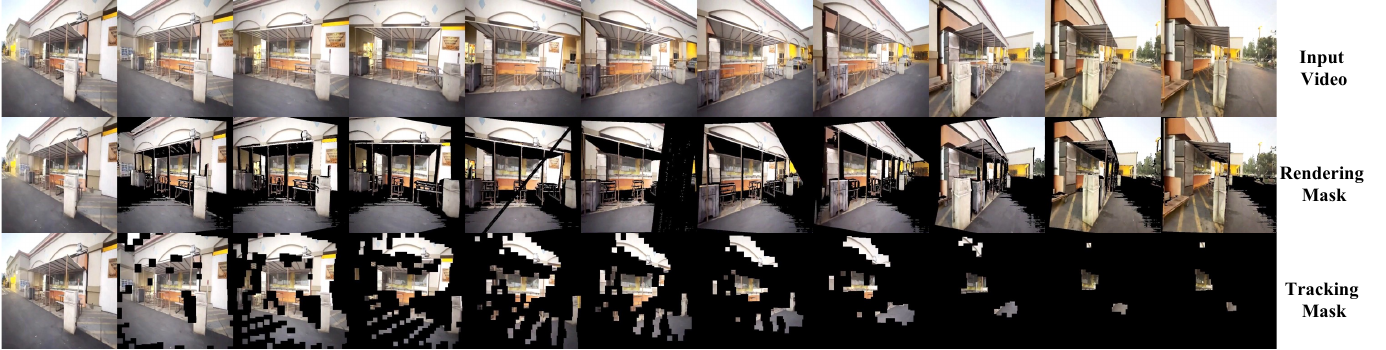}
    \caption{Illustration of our mask generation methods. Top Row: Input Monocular Video; Middle Row: Rendering Mask Generation uses DW-Mesh to simulate occlusions that would occur in novel viewpoints; Bottom Row: Tracking Mask Generation preserves temporal consistency by tracking points across frames and marking consistent occlusion patterns.}
    \label{fig:mask_generation_methods}
\end{figure}

Training video diffusion models for extreme viewpoint synthesis is challenging due to the scarcity of multi-view dynamic video datasets. To address this, we introduce a simulated masking strategy that creates effective training pairs from monocular videos without relying on paired multi-view data. This strategy includes two key components: Rendering Mask Generation and Tracking Mask Generation, as illustrated in Fig.~\ref{fig:mask_generation_methods}.

\paragraph{Rendering Mask Generation.} This component leverages the DW-Mesh to generate occlusion masks for novel viewpoints. For each training video, we:

\phy{1) Employ an off-the-shelf depth estimation method \cite{DepthCrafter} to build DW-Mesh from the input video and remove boundary mesh faces with identical thresholds as defined in Sec. \ref{sec:tex and occ}.}

\phy{2) Render DW-Mesh under the full rotation camera trajectory $\{P_t\}_{t=1}^{T}$ (-90°$\sim$90°) to produce binary visibility masks $\{m_t\}_{t=1}^{T} \in \{0, 1\}^{T \times H \times W}$, with occluded and unseen faces marked as invisible.}

\phy{3) Apply morphological dilation operation on $\{m_t\}_{t=1}^{T}$ to suppress possible noise induced by imprecise depth estimation, obtaining final mask video $V_O$.}

\phy{This process creates realistic occlusion masks that simulate challenging regions during extreme viewpoint synthesis. Applying mask video $V_O$ on corresponding video frames will produce the color video $V_T$, as shown in Fig.~\ref{fig:mask_generation_methods}.} 

\paragraph{Tracking Mask Generation.} To ensure temporal consistency in visibility patterns by tracking points since frame $t$ and marking their surrounding rectangular regions with zero visibility. By maintaining invisibility consistency across frames, this process physically aligns with the actual novel-view video generation, where occluded regions persist across consecutive frames. We employ advanced point tracking model \cite{CoTracker3} to accurately follow points through the video sequence, generating masks that reflect realistic visibility transitions. This technique creates temporally coherent mask sequences that effectively simulate the consistent occlusion patterns observed when viewing scenes from extreme angles, as illustrated in Fig.~\ref{fig:mask_generation_methods}.

\begin{figure}
    \centering
    \includegraphics[width=0.84\linewidth]{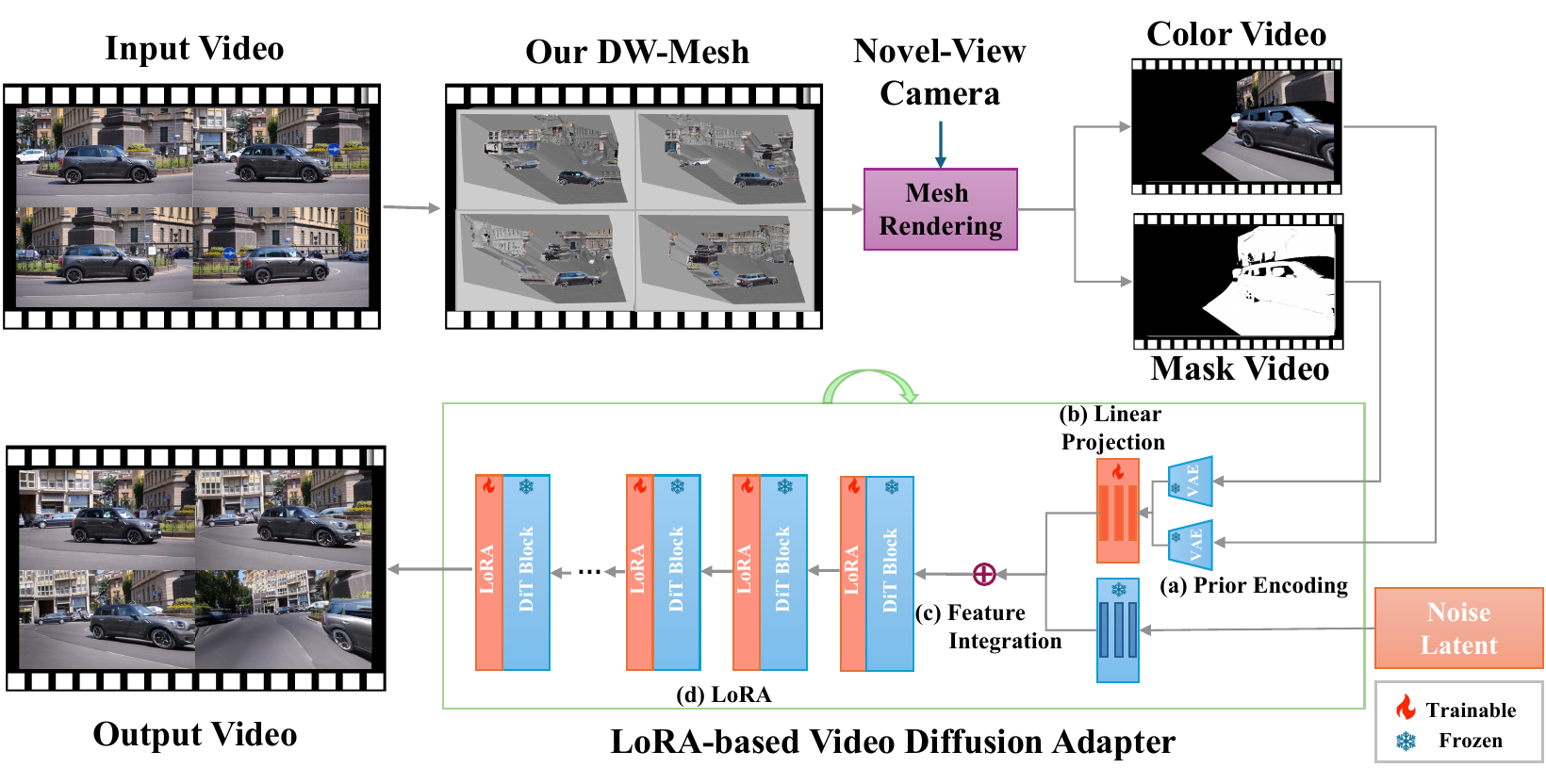}
    \caption{Overview of the EX-4D framework. Our approach transforms monocular videos into extreme viewpoint 4D videos through three key components: (1) Depth Watertight Mesh construction, which explicitly models both visible and occluded regions; (2) Color and mask videos are simulated or rendered for training or inference separately; and (3) a lightweight LoRA-based video diffusion adapter that ensures geometric consistency and temporal coherence in the synthesized 4D videos.}
    \label{fig:overview}
\end{figure}

\subsection{A Lightweight Adapter for Video Diffusion}
To synthesize realistic appearances for novel viewpoints, we build upon a pre-trained image-to-video diffusion model \cite{Wan} and introduce a lightweight adapter designed for mesh-guided video diffusion, \phy{as illustrated in Fig. \ref{fig:overview}}.

Our architecture integrates geometric priors from the DW-Mesh into the video generation process through the following steps:
\textbf{(a) Encoding:} The rendered color and mask videos are passed through a frozen Video VAE encoder to obtain their latent representations.
\textbf{(b) Linear Projection:} These latent features are processed through three linear layers to produce features aligned with the diffusion model's dimensions.
\textbf{(c) Feature Integration:} The projected conditional features are added to the noise latent projection features, seamlessly incorporating geometric priors into the video synthesis pipeline.
\textbf{(d) LoRA-based Adaptation:} To enable efficient training, we utilize a Low-Rank Adaptation (LoRA) \cite{LoRA} approach with a small rank. This allows the pre-trained video diffusion model \cite{Wan} to remain frozen while only updating the adapter parameters.

The training process adheres to the original diffusion model's denoising objective, ensuring rapid convergence and stable performance without requiring extensive computational resources. This design effectively combines geometry-aware conditioning with high-quality video synthesis, maintaining temporal coherence across frames. \phy{The model is trained to predict noise, and the training objective is defined as:}

\begin{equation}
\label{eq:loss}
    \mathcal{L} = \mathbb{E}_{\epsilon,t}[\omega(t) || \epsilon_\theta(z_t,I_1,V_T,V_O,t;\theta)- \epsilon||_2^2]
\end{equation}

\phy{Here $\epsilon$, $z_t$, $\omega(t)$ denotes the ground-truth noise, noisy latents, and training weight at diffusion timestep $t$, respectively. $I_1$ is the first frame of input video. $\epsilon_\theta$ denotes our denoising model with parameters $\theta$.}

\section{Experiments}

\subsection{Experimental Settings}
\paragraph{Datasets.}
For training, since our simulated masking strategy eliminates the need for multi-view videos or camera calibration, we utilize OpenVID \cite{OpenVID}. It is a large-scale monocular video dataset containing over 1 million high-quality monocular videos with rich camera and motion patterns. For evaluation, we construct a testing dataset consisting of 150 in-the-wild web videos, which include diverse dynamic scenes and camera movements.
We evaluate geometric consistency across four increasingly challenging angular ranges: Small (0°$\rightarrow$30°), Large (0°$\rightarrow$60°), Extreme (0°$\rightarrow$90°), and Full (-90°$\sim$90°), using identical camera trajectories for all compared methods.

\paragraph{Metrics.}
We evaluate performance using complementary metrics: Fréchet Inception Distance (FID) \cite{FID} for visual quality assessment, Fréchet Video Distance (FVD) \cite{FVD} for different angular range evaluation, and VBench \cite{VBench} for comprehensive perceptual quality measurement for Full angular Additionally, we conduct a user study to assess generation quality based on human perception. To ensure fair comparisons, all methods are evaluated using identical camera trajectories, and geometry-based methods receive the same depth estimation results as inputs, eliminating any advantage from differing geometric priors.

\paragraph{Baselines.} For comparative evaluation, we involve the state-of-the-art methods capable of camera-controllable video synthesis. Several recent works \cite{Ac3d, CameraCtrl, CamI2V, Syncammaster, ReconX} were excluded as they lack support for arbitrary camera motion control over input videos or reimplementation issues. Our comparison focuses on two classes of approaches: geometry-based video manipulation methods (TrajectoryCrafter \cite{TrajectoryCrafter} and TrajectoryAttention \cite{TrajectoryAttention}) and camera-based conditioning (ReCamMaster \cite{ReCamMaster}). These baselines represent the current leading approaches for novel-view video synthesis, allowing for a comprehensive evaluation of EX-4D's performance in terms of temporal coherence and physical consistency.

\paragraph{Implementation Details.}
For DW-Mesh construction, we use a pre-trained DepthCrafter model \cite{DepthCrafter} for video depth estimation. \phy{We set the depth threshold $\delta_\text{depth} = 0.013(\text{Max}(D_0) - \text{Min}(D_0))$, and $D_\text{max} = 100$ for boundary padding process according to our experiments. Nvdiffrast \cite{nvdiffrast} is adopted as the renderer for both training and validation tasks.} The video diffusion backbone, Wan2.1 \cite{Wan}, remains frozen during training, while our lightweight adapter is configured with LoRA rank of 16. The whole lightweight adapter has around 140M trainable parameters, which only occupies 1\% of the total 14B video diffusion parameters, making it highly efficient. Input videos are resized to $512 \times 512$ resolution with $49$ frames per sequence. The adapter is optimized using the AdamW optimizer with a learning rate of $3 \times 10^{-5}$. Our method is implemented on 32 NVIDIA A100 GPUs (80GB). The training process is completed in only 1 day, and inference generates each video in approximately 4 minutes using 25 denoising steps. Please refer to the supplementary material for additional implementation details and more analysis.

\begin{table}[t]
\centering
\caption{Quantitative comparison of FID and FVD metrics in different viewpoint ranges.}
\label{tab:quantitative}
\resizebox{1.0\linewidth}{!}{
\begin{tabular}{lccccccc}
\toprule
\multirow{2}{*}{\textbf{Method}} & \multicolumn{3}{c}{\textbf{FID$\downarrow$}} & \multicolumn{3}{c}{\textbf{FVD$\downarrow$}} \\
\cmidrule(lr){2-4} \cmidrule(lr){5-7}
& \textbf{Small (0°$\rightarrow$30°)} & \textbf{Large (0°$\rightarrow$60°)} & \textbf{Extreme (0°$\rightarrow$90°)} & \textbf{Small (0°$\rightarrow$30°)} & \textbf{Large (0°$\rightarrow$60°)} & \textbf{Extreme (0°$\rightarrow$90°)} \\
\midrule
\textbf{TrajectoryAttention} \cite{TrajectoryAttention} & 59.86 & 62.69 & \second{62.49} & \second{623.54} & 754.80 & \third{912.14} \\
\textbf{ReCamMaster} \cite{ReCamMaster} & \third{50.88} & \third{56.49} & \third{64.68} & 659.29 &  \second{714.62} & 943.45 \\
\textbf{TrajectoryCrafter} \cite{TrajectoryCrafter} & \second{48.72} & \second{55.24} & 65.33 & \third{633.25} & \third{725.44} & \second{893.80} \\ \hline
\textbf{EX-4D (Ours)} & \best{44.19} & \best{50.30} & \best{55.42} & \best{571.18} & \best{685.39} & \best{823.61} \\
\bottomrule
\end{tabular}
}
\label{tab:fid_fvd}
\end{table}

\begin{table}[t]
    \centering
    \caption{Quantitative comparison between methods in VBench metrics for the Full range (-90°$\rightarrow$90°).}    \label{tab:vbench}
    \resizebox{0.94\linewidth}{!}{
    \begin{tabular}{lccccccc}
    \toprule
    \multirow{2}{*}{\textbf{Method}} & \textbf{Aesthetic} & \textbf{Imaging} & \textbf{Temporal} & \textbf{Motion} & \textbf{Subject} & \textbf{Background} & \textbf{Dynamic} \\
     & \textbf{Quality $\uparrow$} & \textbf{Quality $\uparrow$} & \textbf{Flickering $\uparrow$} & \textbf{Smoothness $\uparrow$} & \textbf{Consistency $\uparrow$} & \textbf{Consistency $\uparrow$} & \textbf{Degree $\uparrow$} \\
    \midrule
    \textbf{TrajectoryAttention} \cite{TrajectoryAttention} & 0.389 & 0.567 & 0.895 & 0.931 & \third{0.834} & 0.846 & 0.923 \\
    \textbf{ReCamMaster} \cite{ReCamMaster} & \third{0.434} & \third{0.582} & \second{0.909} & \best{0.938} & 0.831 & \third{0.849} & \second{0.941} \\
    \textbf{TrajectoryCrafter} \cite{TrajectoryCrafter} & \second{0.447} & \second{0.607} & \third{0.902} & \third{0.928} & \second{0.838} & \second{0.856} & \third{0.936} \\
    \hline
    \textbf{EX-4D (Ours)}  & \best{0.450} & \best{0.631} & \best{0.914} & \second{0.934} & \best{0.846} & \best{0.872} & \best{0.948} \\
    \bottomrule
    \end{tabular}
    }
\end{table}

\subsection{Quantitative Comparison}


The results in Table \ref{tab:fid_fvd} demonstrate that our EX-4D consistently outperforms all baselines across different metrics and viewpoint ranges. For FID scores, our method achieves 44.19, 50.30, and 55.42 for small, large, and extreme viewpoint ranges respectively, showing significant improvements over the second-best method (TrajectoryCrafter with 48.72, 55.24 for small and large ranges, and TrajectoryAttention with 62.49 for extreme angles). Similarly, ours achieves the lowest FVD scores (571.18, 685.39, and 823.61) across all viewpoint ranges, demonstrating superior temporal coherence compared to the baselines. Notably, as the viewpoint angles become more extreme, the performance gap widens, highlighting our method's robustness in handling challenging camera movements.

Table \ref{tab:vbench} shows our method achieves the highest scores on most VBench \cite{VBench} metric, including aesthetic quality (0.450), imaging quality (0.631), and temporal consistency (0.914). Our scores for subject consistency (0.846) and background consistency (0.872) demonstrate the geometric stability of our DW-Mesh representation. While ReCamMaster slightly outperforms us in motion smoothness (0.938 vs. 0.934), our method shows better overall dynamic performance (0.948). Unlike competing methods that degrade significantly with increasing viewpoint angles, our EX-4D maintains consistent quality across extreme camera movements, confirming the effectiveness of our DW-Mesh approach.

\begin{figure}[ht]
    \centering
    \includegraphics[width=0.7\linewidth]{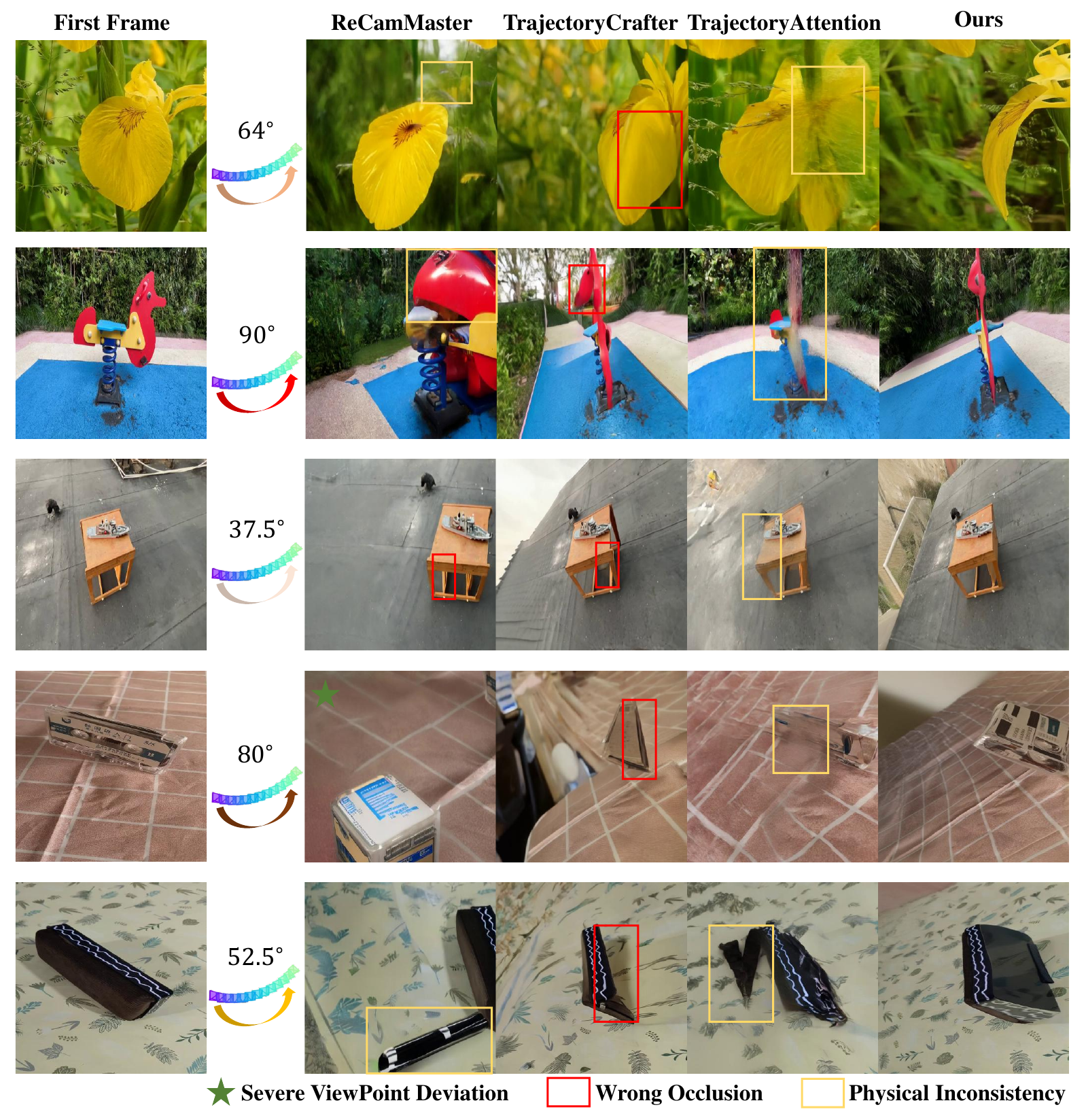}
    \caption{Qualitative comparison of our EX-4D against state-of-the-art approaches under extreme viewpoint. Our approach produces physically consistent videos with effective occlusion handling and temporal coherence. In contrast, baseline methods exhibit artifacts such as \textbf{Physical Inconsistency} and \textbf{Wrong Occlusion} due to their limited ability to model hidden geometry, \phy{or suffer from \textbf{Severe Viewpoint Deviation} in novel scenes outside their training data distribution}.\vspace{-0.2in}}
    \label{fig:qualitative_comparison}
\end{figure}

\subsection{Qualitative Comparison}

We present a qualitative comparison between our EX-4D approach and state-of-the-art methods in Fig. \ref{fig:qualitative_comparison}. Existing geometry-based approaches (TrajectoryCrafter and TrajectoryAttention) show significant limitations with extreme viewpoints, producing ghosting artifacts, geometric distortions, and blurring in occluded regions due to their inability to properly model hidden surfaces. ReCamMaster, despite being trained on expensive multi-view datasets, exhibits inconsistent object boundaries and struggles with extreme camera trajectories, often generating structurally implausible content for regions that become visible during viewpoint changes. In contrast, our method produces physically consistent videos with effective occlusion handling, maintaining object shapes, surface details, and temporal coherence even under extreme camera movements. These results demonstrate the effectiveness of our DW-Mesh representation, which explicitly models both visible and hidden regions to ensure physical plausibility across arbitrary viewpoints.

\subsection{User Study}
\begin{figure}[t]
    \centering
    \includegraphics[width=0.5\linewidth]{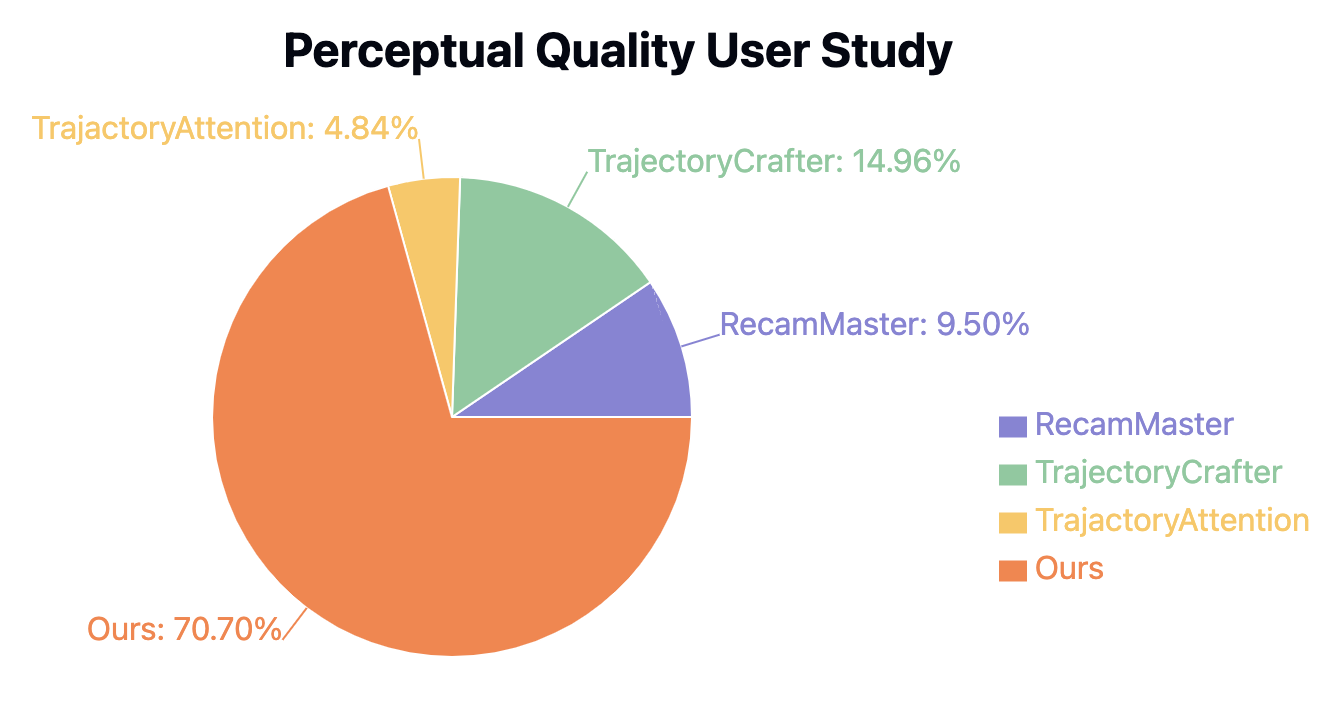}
    \caption{User study results comparing our EX-4D method against baselines. Participants evaluated videos based on quality of physical consistency and extreme viewpoint, with our approach receiving significantly higher preference ratings.}
    \label{fig:user_study}
\end{figure}

We conducted a user study to evaluate perceptual qualities not fully captured by metrics like FID and FVD. Fifty participants viewed 12 randomly selected video sets from our testing dataset, comparing our method with baselines \cite{ReCamMaster,TrajectoryAttention,TrajectoryCrafter}. Participants selected which method best maintained physical consistency and produced convincing extreme viewpoints. As shown in Fig. \ref{fig:user_study}, our method was strongly preferred with 70.70\% of votes, compared to TrajectoryCrafter (14.96\%), ReCamMaster (9.50\%), and TrajectoryAttention (4.84\%). This preference highlights our method's ability to avoid common artifacts like warping distortions and temporal flickering. By modeling occluded regions with DW-Mesh, our approach ensures natural transitions as objects move in and out of view, aligning with human perception.

\begin{table}[t]
\centering
\caption{Ablation study on the components of EX-4D in \textbf{Extreme Viewpoint (0°$\rightarrow$90°)}.}
\label{tab:ablation}
\definecolor{lightcoral}{rgb}{1.0, 0.7, 0.7}
\definecolor{lightgreen}{rgb}{0.7, 0.9, 0.7}
\resizebox{0.8\linewidth}{!}{
\begin{tabular}{lll}
\toprule
Variant & FID$\downarrow$ & FVD$\downarrow$ \\
\midrule
Full Method (DW-Mesh + LoRA Rank 16 + Wan2.1) & \textbf{55.42} & \textbf{823.61} \\ \hline
w/ ControlNet \cite{ControlNet} & Out Of Memory  & Out Of Memory \\
w/ CogVideoX Backbone & 59.76 \textcolor{lightcoral}{(worse 7.831\%)} & 867.25 \textcolor{lightcoral}{(worse 5.299\%)} \\
w/o DW-Mesh & 74.31 \textcolor{lightcoral}{(worse 34.085\%)} & 1103.21 \textcolor{lightcoral}{(worse 33.948\%)} \\
w/ Random Masks & 69.36 \textcolor{lightcoral}{(worse 25.153\%)} & 993.64 \textcolor{lightcoral}{(worse 20.644\%)} \\
w/o Rendering Masks & 63.35 \textcolor{lightcoral}{(worse 14.309\%)} & 972.93 \textcolor{lightcoral}{(worse 18.130\%)} \\
w/o Tracking Masks & 60.24 \textcolor{lightcoral}{(worse 8.697\%)} & 924.47 \textcolor{lightcoral}{(worse 12.246\%)} \\
w/ LoRA Rank 64 & 53.68 \textcolor{lightgreen}{(better 3.140\%)} & 802.47 \textcolor{lightgreen}{(better 2.567\%)} \\
\bottomrule
\end{tabular}
}
\end{table}

\subsection{Ablation Study}
To assess the impact of each component in our framework, we performed an ablation study by modifying or removing key elements. The results, shown in Table \ref{tab:ablation}, highlight the following: Replacing our lightweight LoRA-based adapter with ControlNet \cite{ControlNet} causes an out-of-memory error on 80GB GPUs, emphasizing our approach's efficiency. Removing the DW-Mesh leads to the largest performance drop, with FID increasing by 34.1\% to 74.31 and FVD worsening by 33.9\% to 1103.21, underscoring its importance for handling occlusions and ensuring consistency. Using random masks instead of structured ones significantly degrades performance (FID 69.36, FVD 993.64), showing the value of meaningful geometric guidance. Both masking strategies are crucial—removing rendering masks increases FID by 14.3\% and FVD by 18.1\%, while removing tracking masks raises FID by 8.7\% and FVD by 12.2\%. Switching from Wan2.1 to CogVideoX reduces performance by 7.8\% on FID and 5.3\% on FVD. Increasing the LoRA rank from 16 to 64 yields only slight improvements (3.1\% on FID, 2.6\% on FVD), indicating our lightweight adapter is already effective and efficient. Additional details are in the supplementary material.

\section{Conclusion}
We introduced EX-4D, a framework for generating high-quality 4D videos from monocular input under extreme viewpoints. Our DW-Mesh ensures geometric consistency by modeling both visible and occluded regions, while our masking strategy avoids the need for multi-view training data. A lightweight LoRA-based adapter synthesizes coherent videos efficiently. Experiments confirm EX-4D outperforms existing methods, especially at extreme angles. \textbf{Limitations.} EX-4D relies on depth estimation, which may struggle with reflective or transparent surfaces, and requires significant computation for high-resolution videos. Future work will focus on improving depth robustness and computational efficiency. (Limitations and Failure cases are discussed in supplemental material.)

{
    \small
    \bibliographystyle{ieeenat_fullname}
    \bibliography{main}
}

\newpage

\newpage
\appendix 
\section{Supplementary Material}

\subsection{Network Structure}
The EX-4D Adapter consists of four main modules: Prior Encoding, Linear Projection, Feature Integration, and LoRA. Below, we provide detailed descriptions of each module:

\textbf{Prior Encoding} leverages a frozen Video VAE encoder from Wan Text-to-Vodeo model \cite{Wan} to extract compact latent representations from both the input color video and the corresponding mask video. Specifically, given input sequences of shape $\mathbb{R}^{49\times512\times512}$, the VAE encodes each into latent tensors of shape $\mathbb{R}^{7\times64\times64}$, where $49$ is the number of frames and $512\times512$ is the spatial resolution. This encoding preserves essential spatiotemporal information while significantly reducing dimensionality, enabling efficient downstream processing. The encoded latents from the color and mask videos are then concatenated along the channel dimension to form a unified geometric prior, which is subsequently fed into the linear projection module for further feature transformation.

\textbf{Linear Projection} is implemented as a sequence of $1\times1\times1$ Conv3d layers followed by a final Conv3d layer with kernel size $(1,2,2)$ and stride $(1,2,2)$. The concatenated latent features from the prior encoding stage are first projected to a higher-dimensional hidden space using the $1\times1\times1$ convolutions with SiLU activations. The final Conv3d layer then downsamples the spatial dimensions to produce patch embeddings that match the expected input shape of the diffusion model. This design ensures efficient channel mixing and spatial alignment between the geometric priors and the video diffusion backbone. 

The following code snippet illustrates the implementation of both prior encoding and linear projection layer within the EX-4D adapter:
\begin{lstlisting}[language=Python, caption={EX-4D Adapter: Prior Encoding and Linear Projection.}, label={lst:adapter_structure}, xleftmargin=1.5em]
import torch
import torch.nn as nn

class PriorEncoding(nn.Module):
    """
    A VAE model for encoding camera information and video features.
    """

    def __init__(
        self,
        in_channels: int = 16,
        hidden_channels: int = 1024,
        out_channels: int = 5120,
    ) -> None:
        super().__init__()

        self.latent_encoder = torch.nn.Sequential(
            torch.nn.Conv3d(in_channels * 2, hidden_channels, kernel_size=1, stride=1, padding=0),
            torch.nn.SiLU(),
            torch.nn.Conv3d(hidden_channels, hidden_channels, kernel_size=1, stride=1, padding=0),
            torch.nn.SiLU(),
            torch.nn.Conv3d(hidden_channels, hidden_channels, kernel_size=1, stride=1, padding=0)
        )
        self.latent_patch_embedding = torch.nn.Conv3d(hidden_channels, out_channels, kernel_size=(1, 2, 2), stride=(1, 2, 2))
        nn.init.zeros_(self.latent_patch_embedding.weight)
        nn.init.zeros_(self.latent_patch_embedding.bias)

    def _set_gradient_checkpointing(self, module, value=False):
        if isinstance(module, nn.Module):
            module.gradient_checkpointing = value

    def forward(self, video, mask, vae) -> torch.Tensor:
        with torch.no_grad():
            video = vae.encode(video, device=video.device)
            mask = vae.encode(mask * 2 - 1, device=mask.device)
        latent = torch.cat([video, mask], dim=1)
        latent = self.latent_encoder(latent)
        latent = self.latent_patch_embedding(latent)
        return latent

def prepare_camera_embeds(
    prior_encoding,
    vae,
    video,
    mask=None,
) -> torch.Tensor:
    prior_latent = prior_encoding(video, mask, vae)
    return prior_latent

\end{lstlisting}

\textbf{Feature Integration} fuses the projected geometric priors with the noise latent features used in the diffusion process. The integration is performed by element-wise addition, allowing the model to condition the generation process on both the appearance and occlusion information encoded in the priors. This design enables the adapter to inject geometric consistency and mask-aware guidance into the video synthesis pipeline.

\begin{lstlisting}[language=Python, caption={EX-4D Adapter: Feature Integration.}, label={lst:feature_intergration}, xleftmargin=1.5em]
import torch
import torch.nn as nn
% === EX-4D: Prior Encoding ===
x = self.patch_embedding(noise_latent)
prior_latent = prior_encoding(video, mask, vae)

% === Start: EX-4D Adapter: Feature Integration ===
x = self.patch_embedding(noise_latent)
x = x + prior_latent
% === End: EX-4D Adapter: Feature Integration ===

% === EX-4D: Diffusion Transformer ===
x = self.transformer(x, context, time_embedding)
\end{lstlisting}

\textbf{LoRA (Low-Rank Adaptation)} is employed to enable efficient fine-tuning of the adapter with minimal trainable parameters. In our implementation, LoRA layers are applied to the following modules: \texttt{q}, \texttt{k}, \texttt{v}, \texttt{o}, \texttt{ffn.0}, and \texttt{ffn.2} within each attention block of the video diffusion backbone. The LoRA module introduces low-rank updates to these linear projection weights, allowing the adapter to adapt to new tasks or domains without updating the full set of backbone parameters. This approach significantly reduces memory and computational consumption, making the EX-4D Adapter lightweight and scalable for large-scale video generation tasks.

Together, these modules enable the EX-4D Adapter to effectively incorporate geometric priors and mask information into the video diffusion process, resulting in high-quality, physically consistent, and temporally coherent 4D video synthesis under extreme viewpoints.

\subsection{LoRA Integration in Video Diffusion Models}
We employ Low-Rank Adaptation (LoRA) to efficiently fine-tune our video diffusion backbone. The following Python function demonstrates how LoRA modules are injected into a model, targeting specific layers such as attention projections and feed-forward blocks. This approach enables parameter-efficient adaptation by updating only a small subset of weights.

\begin{lstlisting}[language=Python, caption={EX-4D Adapter: Feature Integration module.}, label={lst:lora_integration}, xleftmargin=1.5em]
def add_lora_to_model(self, model, lora_rank=16, lora_alpha=16, lora_target_modules="q,k,v,o,ffn.0,ffn.2", init_lora_weights="kaiming", pretrained_path=None, state_dict_converter=None):
    # Add LoRA to UNet
    self.lora_alpha = lora_alpha
    if init_lora_weights == "kaiming":
        init_lora_weights = True
        
    lora_config = LoraConfig(
        r=lora_rank,
        lora_alpha=lora_alpha,
        init_lora_weights=init_lora_weights,
        target_modules=lora_target_modules.split(","),
    )
    model = inject_adapter_in_model(lora_config, model)
    for param in model.parameters():
        # Upcast LoRA parameters into fp32
        if param.requires_grad:
            param.data = param.to(torch.float32)
            
    # Lora pretrained lora weights
    if pretrained_path is not None:
        state_dict = load_state_dict(pretrained_path)
        if state_dict_converter is not None:
            state_dict = state_dict_converter(state_dict)
        missing_keys, unexpected_keys = model.load_state_dict(state_dict, strict=False)
        all_keys = [i for i, _ in model.named_parameters()]
        num_updated_keys = len(all_keys) - len(missing_keys)
        num_unexpected_keys = len(unexpected_keys)
        print(f"LORA: {num_updated_keys} parameters are loaded from {pretrained_path}. {num_unexpected_keys} parameters are unexpected.")
\end{lstlisting}

This function configures and injects LoRA modules into the specified target layers, optionally loading pretrained LoRA weights. It ensures all trainable parameters are in \texttt{float16} for numerical stability. This design allows for scalable and memory-efficient adaptation of large video diffusion models.

\begin{figure}[ht]
    \centering
    \begin{subfigure}[b]{0.8\linewidth}
        \includegraphics[width=\linewidth]{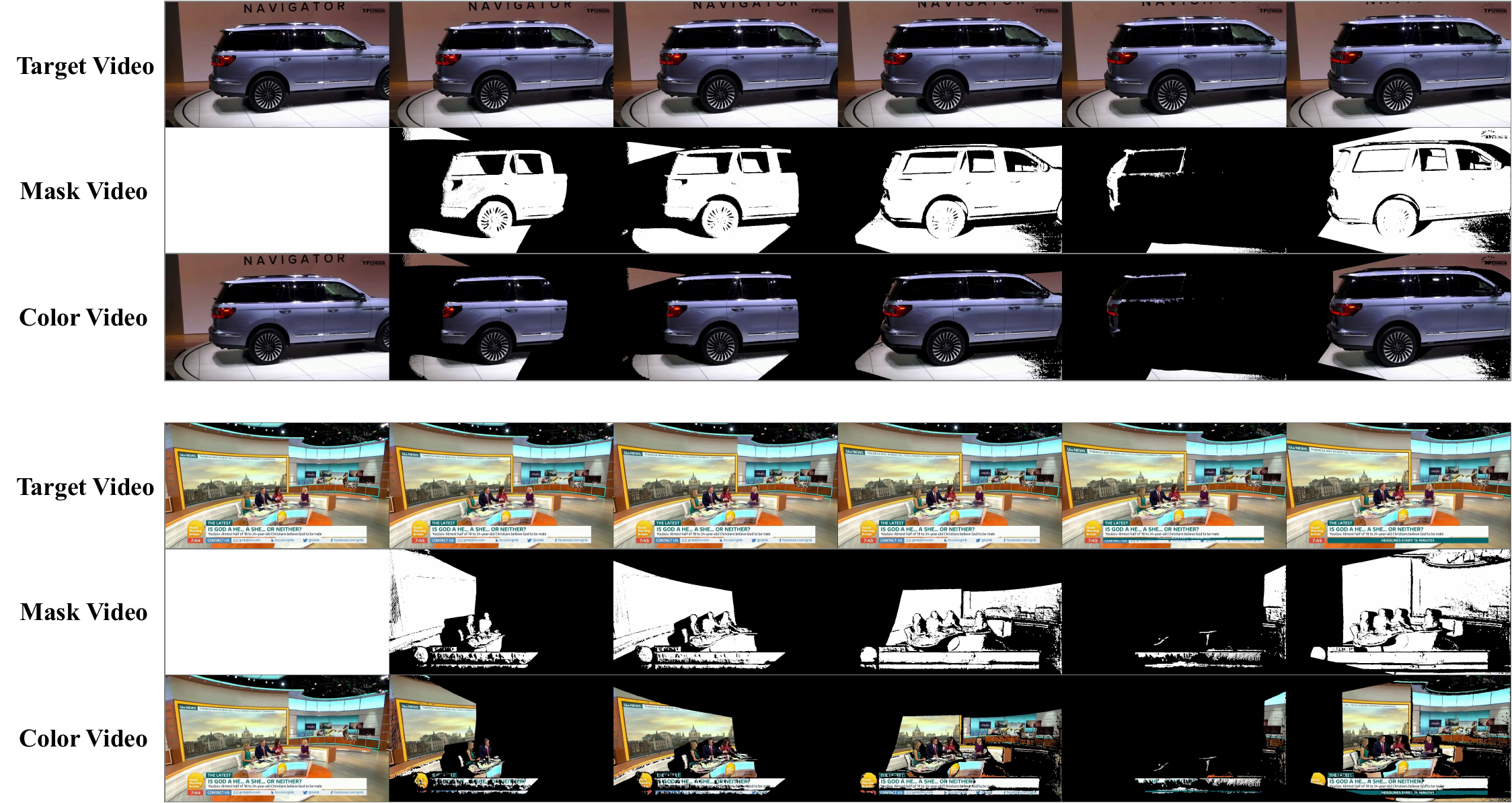}
        \caption{Rendering Mask Generation: Using DW-Mesh to simulate occlusions from novel viewpoints.}
        \label{fig:rendering_mask_details}
    \end{subfigure}
    \vspace{0.2in}
    \begin{subfigure}[b]{0.8\linewidth}
        \includegraphics[width=\linewidth]{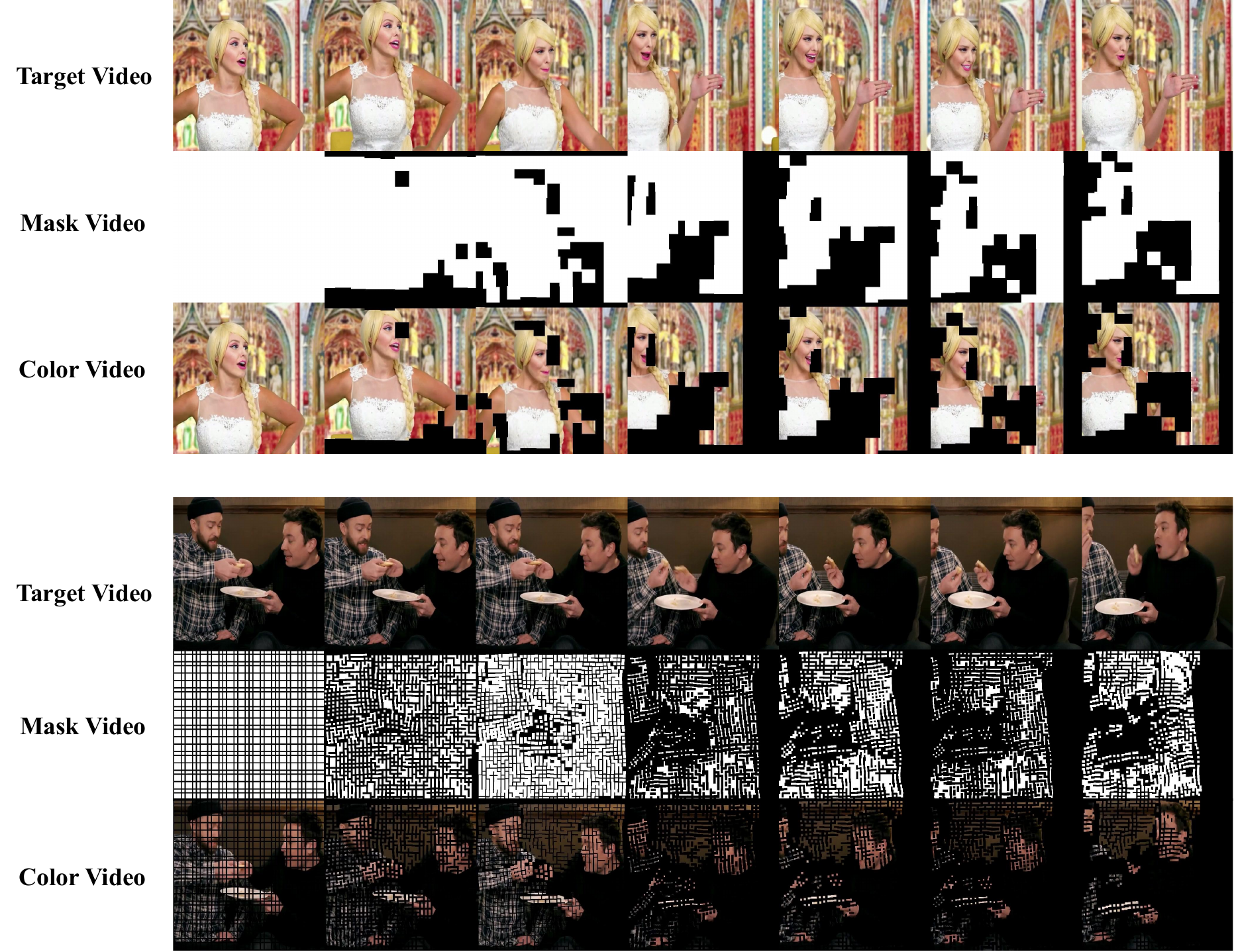}
        \caption{Tracking Mask Generation: Preserving temporal consistency through point tracking across frames.}
        \label{fig:tracking_mask_details}
    \end{subfigure}
    \caption{Detailed visualization of our mask generation methods. (a) Rendering masks are created by simulating novel viewpoint occlusions using the DW-Mesh representation. (b) Tracking masks ensure temporally consistent occlusion patterns by tracking points across consecutive frames.}
    \label{fig:mask_generation_details}
\end{figure}

\subsection{Details about Mask Generation}
Fig.~\ref{fig:mask_generation_details} illustrates more examples about our rendering and tracking mask approaches. 

Rendering mask generation relies on uniform sampling of diverse viewpoint angles across the full -90° to 90° range, ensuring comprehensive coverage of potential camera positions during inference. This technique leverages the DW-Mesh representation to simulate realistic occlusions that would occur when viewing the scene from novel perspectives. \phy{To ensure the generation of realistic occlusion masks, we enforce adjacent faces $\{(i,j), (i+1,j), (i,j+1)\}$ and $\{(i+1,j), (i,j+1), (i+1,j+1)\}$ must be either simultaneously occluded or unoccluded. Subsequently, We apply morphological dilation operation with the kernel size of $5 \times 5$ on the binary mask. This process effectively removes isolated noise pixels while preserving the structural integrity of major occlusion regions, ensuring smooth and continuous occlusion boundaries.}

\begin{figure}[ht]
    \centering
    \includegraphics[width=0.85\linewidth]{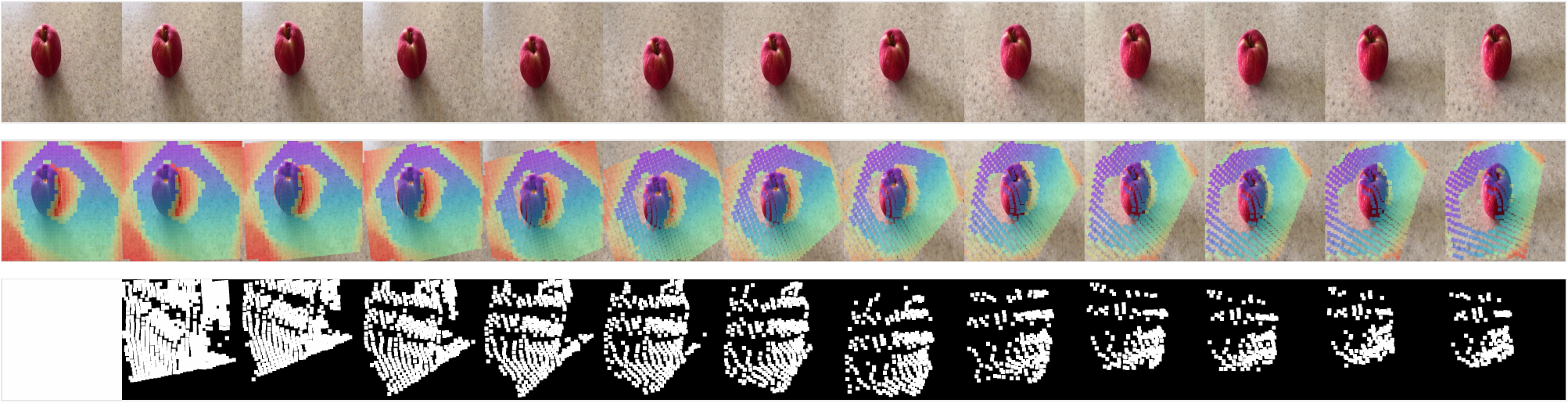}
    \caption{Principle of tracking mask generation. Points are tracked across frames to create consistent occlusion patterns, ensuring temporal coherence. Different colors represent corresponding tracked points between frames, helping maintain consistent visibility relationships during motion.}
    \label{fig:tracking_mask_principle}
\end{figure}

The tracking mask approach establishes a grid of 10-50 points per frame, with grid size randomly selected for each training instance to ensure model learning from varied point distributions. We maintain balance between spatial coverage and computational efficiency by adjusting density based on scene complexity. An off-the-shelf tracker \cite{CoTracker3} follows points across consecutive frames, preserving consistent visibility patterns to simulate temporal occlusion effects. The principle of tracking mask generation is illustrated in Fig.~\ref{fig:tracking_mask_principle}.

Additional video augmentation techniques enhance training diversity. Our smooth cropping procedure operates in both horizontal and vertical directions, using crop window sizes of 85-95\% of the original frame. Rather than static crops, we generate smooth trajectories following Bezier curves with controlled acceleration and deceleration. This approach introduces viewpoint variations without requiring explicit 3D understanding, improving the model's ability to generalize to diverse camera movements.

\subsection{Failure Cases}
Despite EX-4D's effectiveness for extreme viewpoint synthesis, several challenging scenarios can lead to suboptimal results:

\begin{figure}[t]
    \centering
    \begin{subfigure}[b]{0.8\linewidth}
        \includegraphics[width=\linewidth]{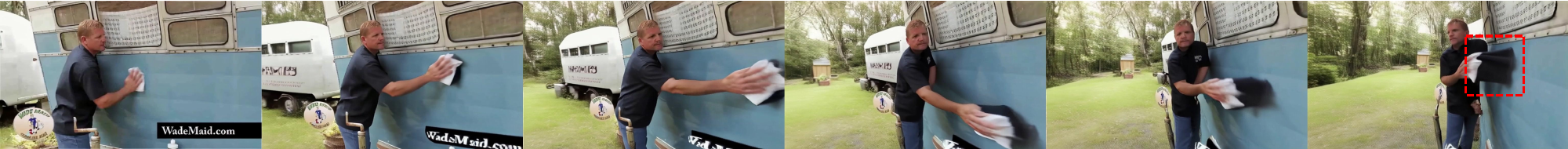}
        \caption{Failure due to inaccurate depth estimation: incorrect geometry leads to distorted occlusion boundaries.}
        \label{fig:failure_depth}
    \end{subfigure}
    \begin{subfigure}[b]{0.8\linewidth}
        \includegraphics[width=\linewidth]{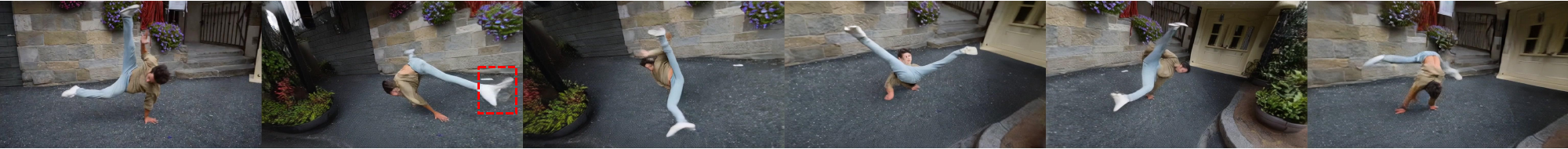}
        \caption{Failure on fine/thin structures: mesh oversmoothing or missing thin objects causes loss of detail or floating artifacts.}
        \label{fig:failure_thin}
    \end{subfigure}
    \caption{Representative failure cases of EX-4D. (a) Depth estimation errors causing visible distortions in novel views; (b) Fine structure handling limitations where thin objects are lost or misrepresented.}
\end{figure}

\paragraph{Depth Estimation Limitations.} Our framework relies heavily on monocular depth estimation quality. When depth maps contain errors due to challenging scenes (reflective surfaces, complex lighting, rapid motion), the resulting DW-Mesh may exhibit geometric inaccuracy. As shown in Fig.~\ref{fig:failure_depth}, these inaccuracies can propagate to synthesized views, causing visible distortions or incorrect occlusion boundaries.

\paragraph{Fine Geometric Detail Preservation.} The watertight mesh construction process may struggle with very thin structures or fine details. Features like wires, fences, or small protruding elements might be oversmoothed or entirely missing in the reconstructed geometry. Fig.~\ref{fig:failure_thin} demonstrates how this limitation can result in loss of detail or floating artifacts in rendered outputs.

Future work could address these limitations through multi-frame depth consistency enforcement, uncertainty-aware mesh construction, or incorporating semantic cues to better preserve important structural elements. Exploring neural mesh refinement techniques may also improve handling of ambiguous regions while maintaining the computational efficiency of our approach.

\subsection{User Study Settings}
We conducted a comprehensive user study to evaluate the perceptual quality of our method compared to baseline approaches. The study involved 50 participants evaluating 12 randomly selected video sequences from our test dataset. Each video sequence contained results from our EX-4D method and all three baseline approaches (ReCamMaster, TrajectoryCrafter, and TrajectoryAttention).

\begin{figure}[ht]
    \centering
    \includegraphics[width=0.8\linewidth]{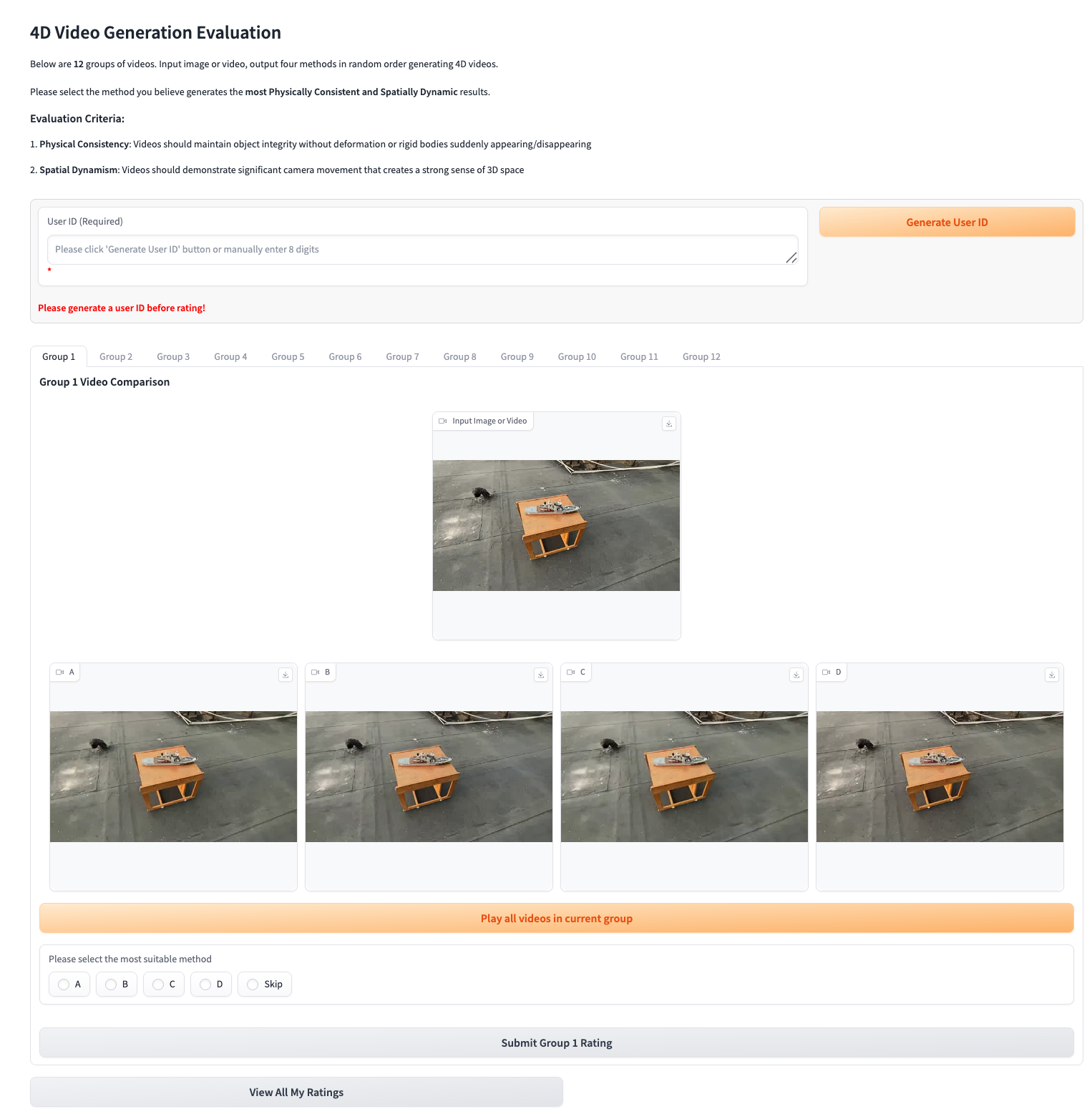}
    \caption{User study interface. Participants were presented with four methods (labeled A-D in randomized order) and asked to select the one that produced the most physically consistent and convincing extreme viewpoint videos. The interface allowed for multiple viewings before selection to ensure informed comparisons.}
    \label{fig:user_study_interface}
\end{figure}

Participants were asked to select which method produced the most visually compelling results based on two key criteria: physical consistency (maintaining object integrity without unrealistic deformations) and extreme viewpoint quality (demonstrating significant camera movement with a strong sense of 3D space). As shown in Fig.\ref{fig:user_study_interface}, the study interface presented videos in randomized order (labeled as Methods A-D) to avoid position bias.

To ensure reliable results, we included attention check questions and allowed participants to replay videos multiple times before making selections. The results, as presented in Fig.~\ref{fig:user_study}, showed a strong preference for our method, with 70.70\% of participants selecting EX-4D as producing the most physically consistent and convincing extreme viewpoint videos.

\subsection{More Visualization}
\begin{figure}[ht]
    \centering
    \begin{subfigure}{0.8\linewidth}
        \includegraphics[width=\linewidth]{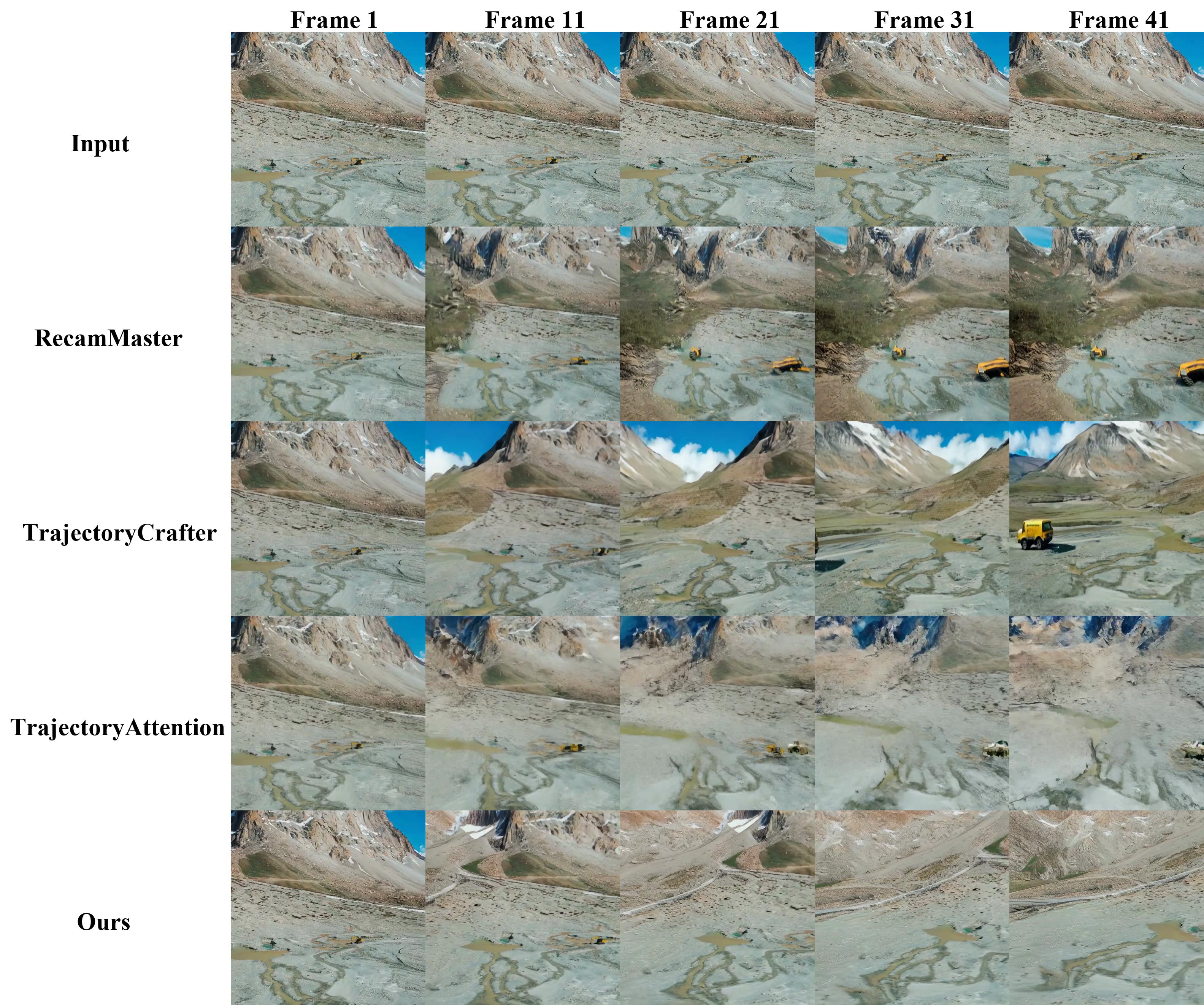}
    \end{subfigure}
    \begin{subfigure}{0.8\linewidth}
        \includegraphics[width=\linewidth]{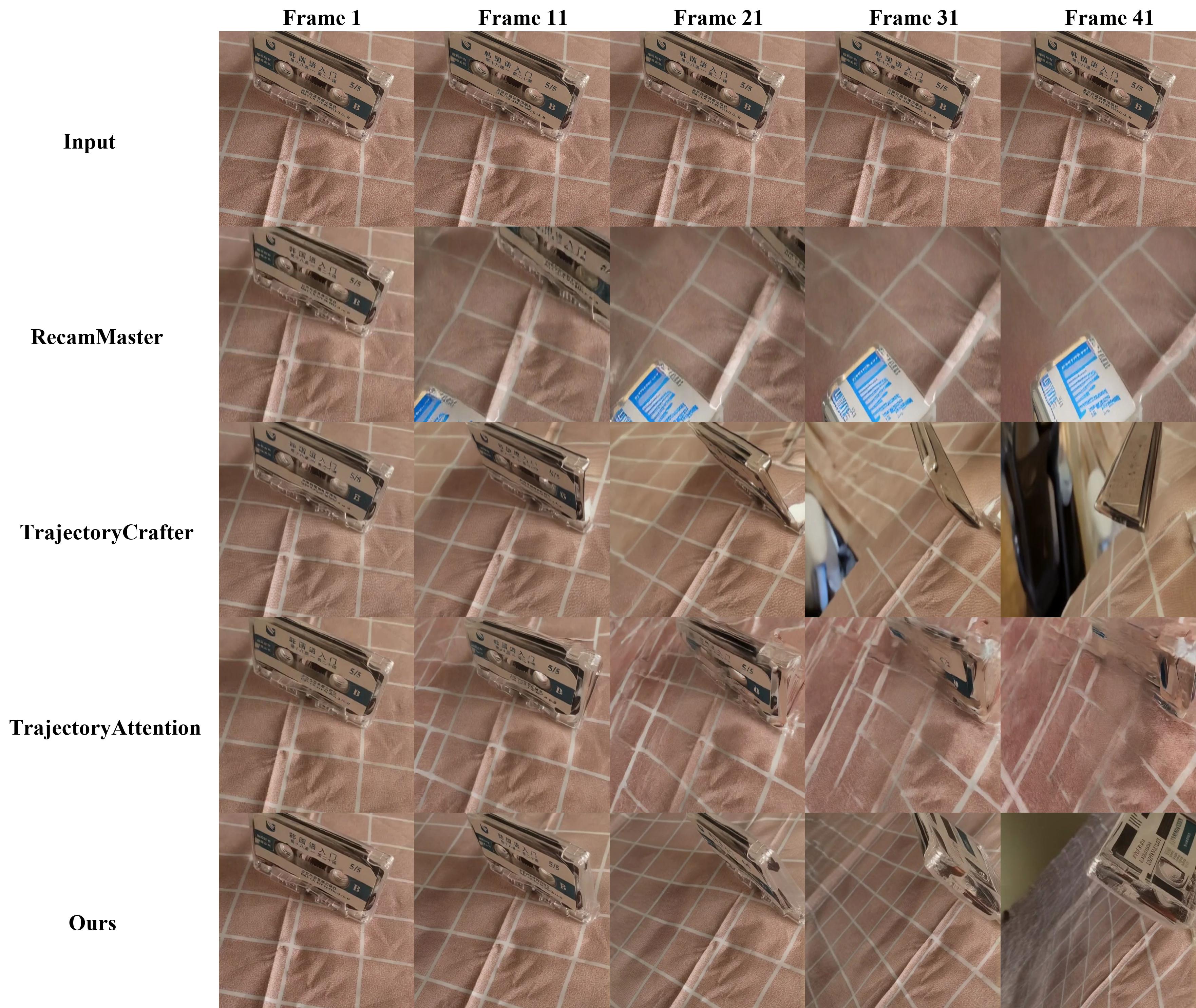}
    \end{subfigure}
    \caption{Comparison of EX-4D with state-of-the-art methods.}
    \label{fig:extra_results1}
\end{figure}

\begin{figure}[ht]
    \centering
    \begin{subfigure}{0.9\linewidth}
        \includegraphics[width=\linewidth]{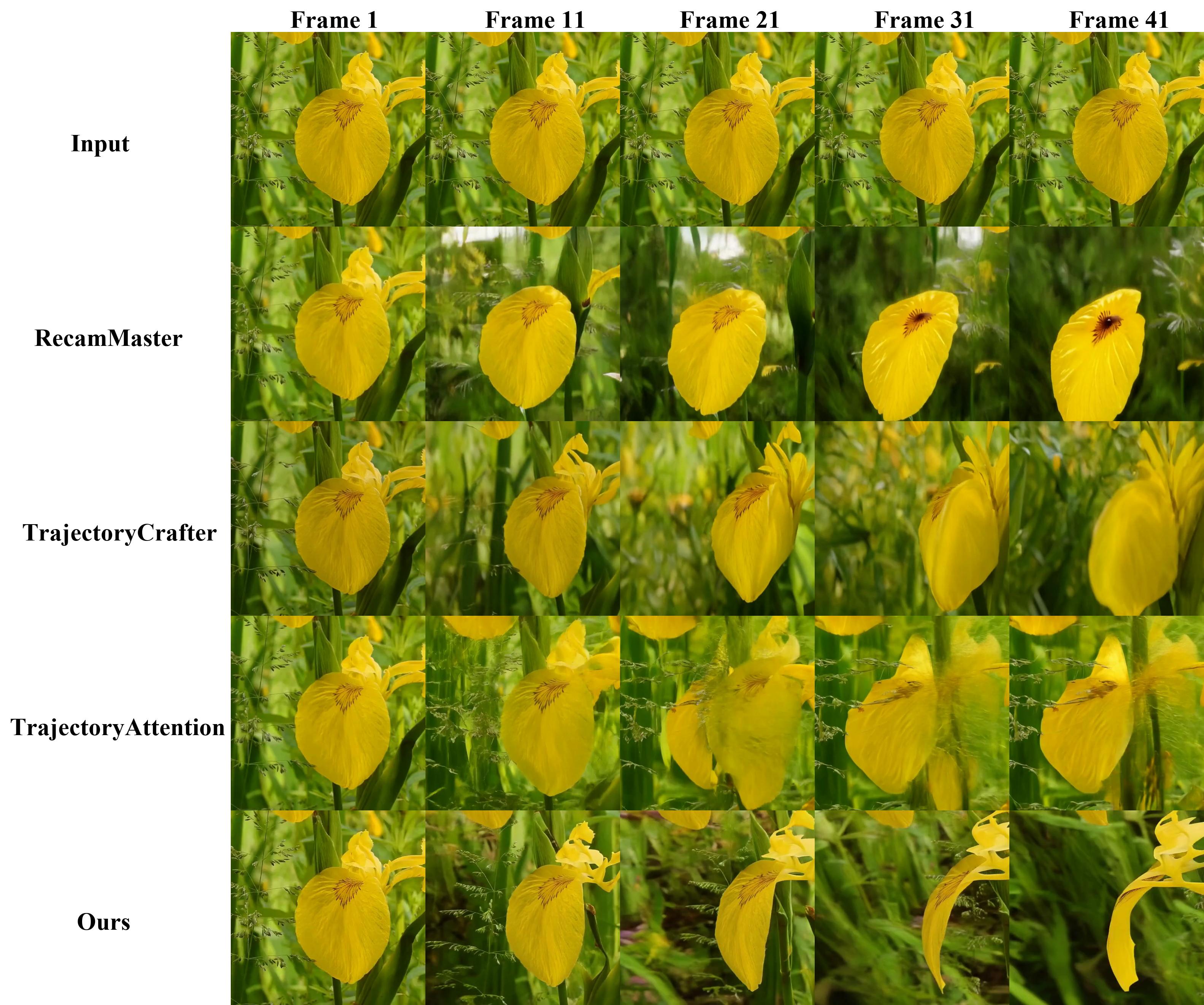}
    \end{subfigure}
    \begin{subfigure}{0.9\linewidth}
        \includegraphics[width=\linewidth]{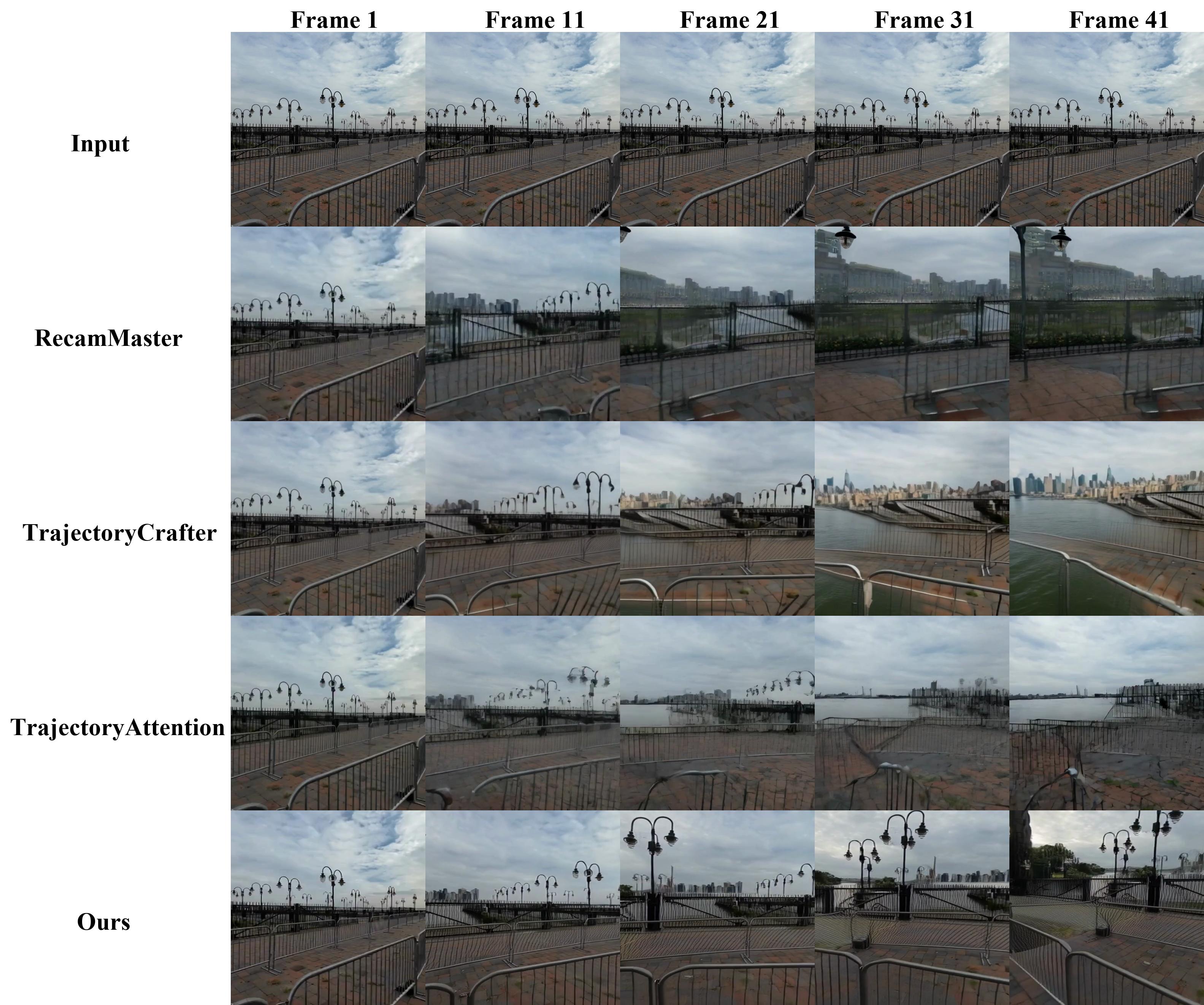}
    \end{subfigure}
    \caption{Comparison of EX-4D with state-of-the-art methods.}
    \label{fig:extra_results2}
\end{figure}

\begin{figure}[ht]
    \centering
    \begin{subfigure}{0.9\linewidth}
        \includegraphics[width=\linewidth]{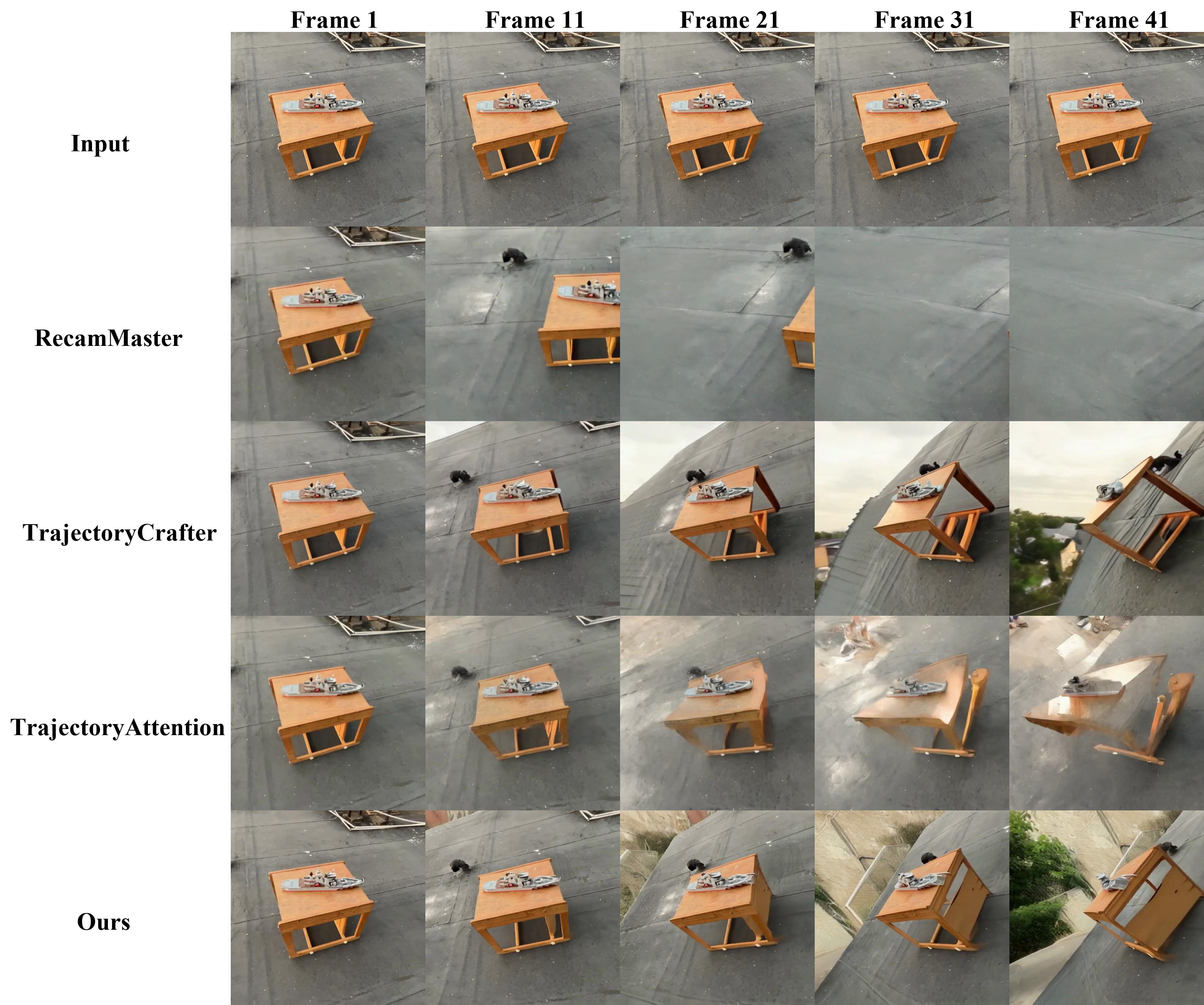}
    \end{subfigure}
    \begin{subfigure}{0.9\linewidth}
        \includegraphics[width=\linewidth]{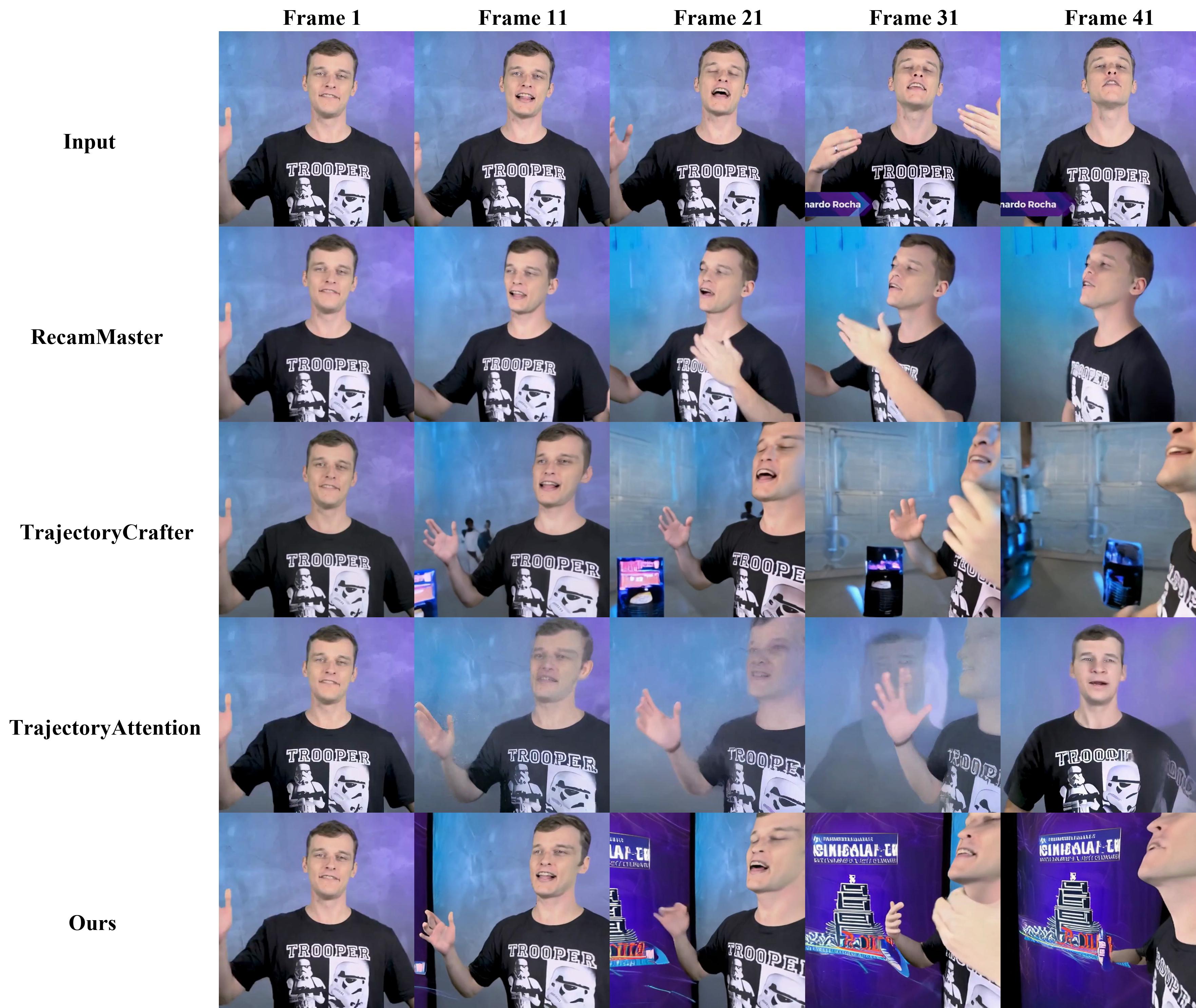}
    \end{subfigure}
    \caption{Comparison of EX-4D with state-of-the-art methods.}
    \label{fig:extra_results3}
\end{figure}

\begin{figure}[ht]
    \centering
    \begin{subfigure}{0.9\linewidth}
        \includegraphics[width=\linewidth]{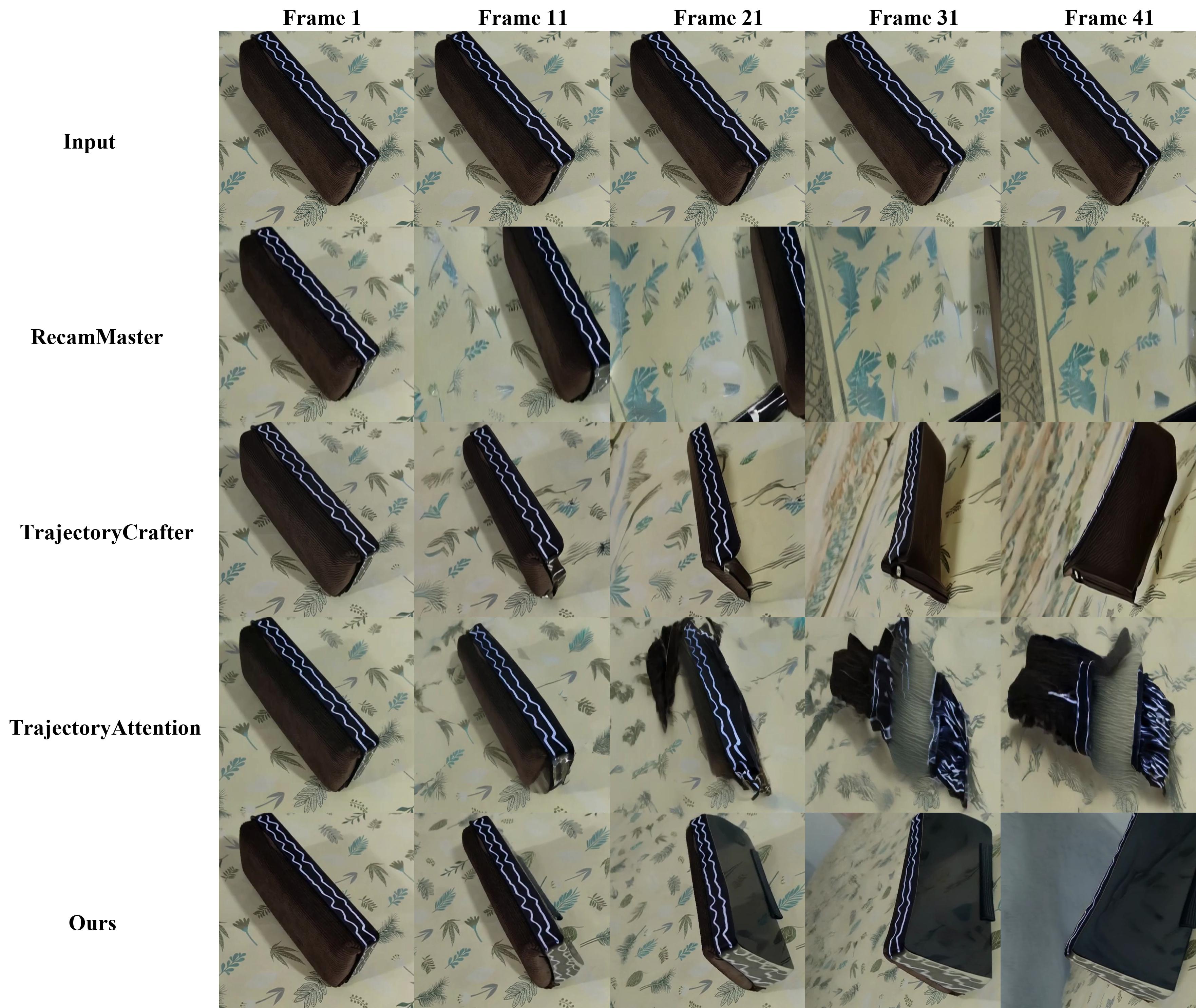}
    \end{subfigure}
    \begin{subfigure}{0.9\linewidth}
        \includegraphics[width=\linewidth]{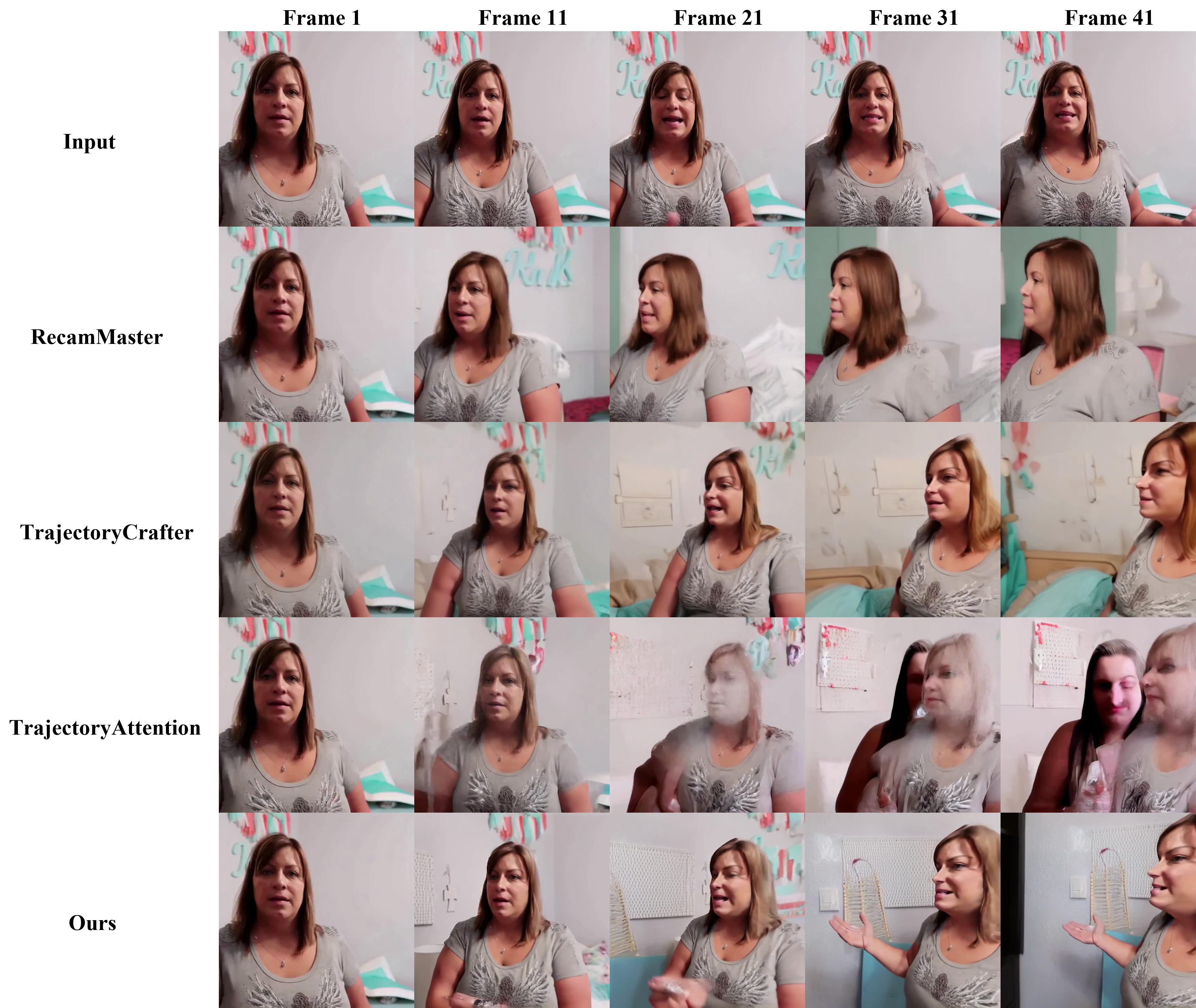}
    \end{subfigure}
    \caption{Comparison of EX-4D with state-of-the-art methods.}
    \label{fig:extra_results4}
\end{figure}

\begin{figure}[ht]
    \centering
    \begin{subfigure}{0.9\linewidth}
        \includegraphics[width=\linewidth]{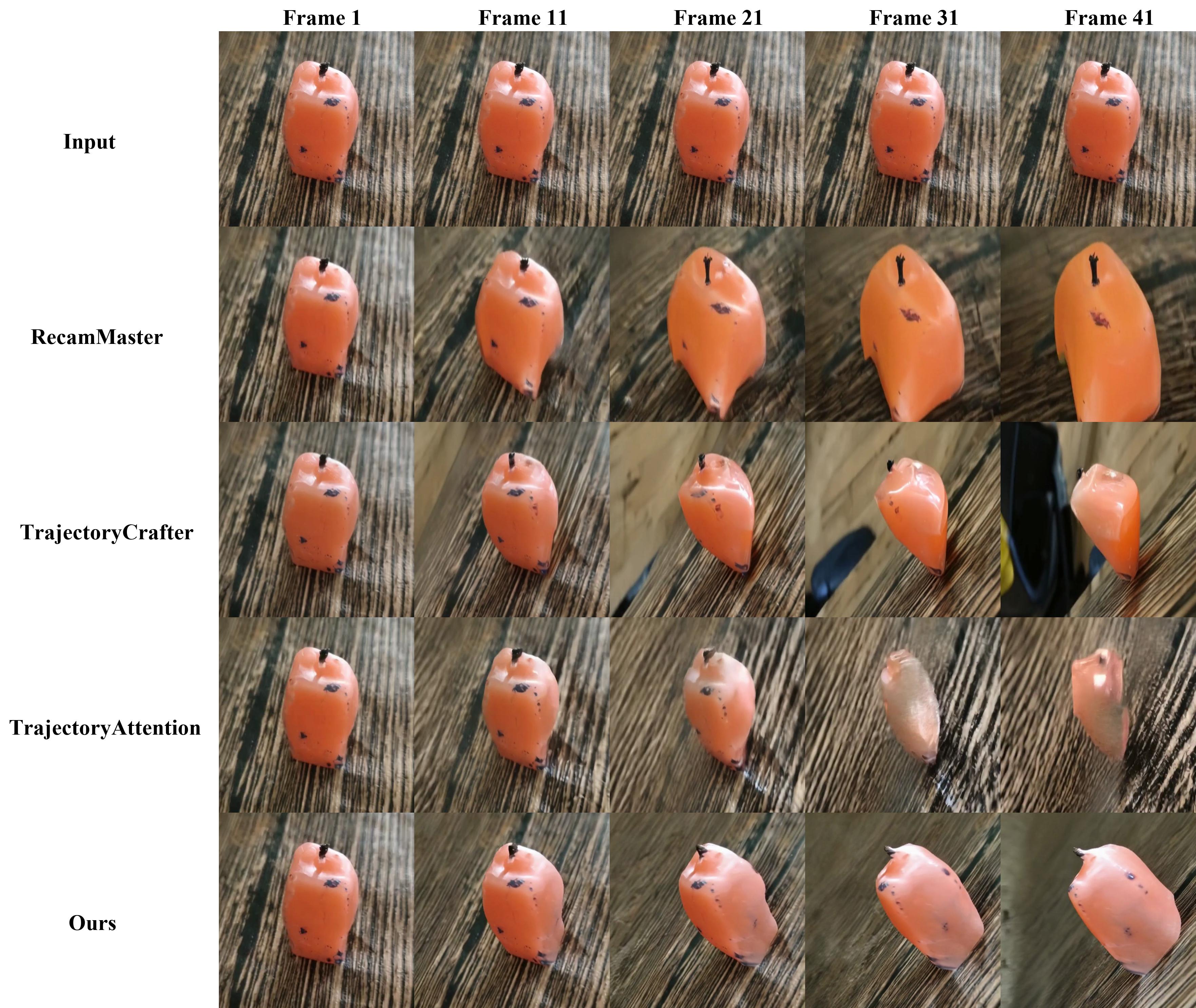}
    \end{subfigure}
    \begin{subfigure}{0.9\linewidth}
        \includegraphics[width=\linewidth]{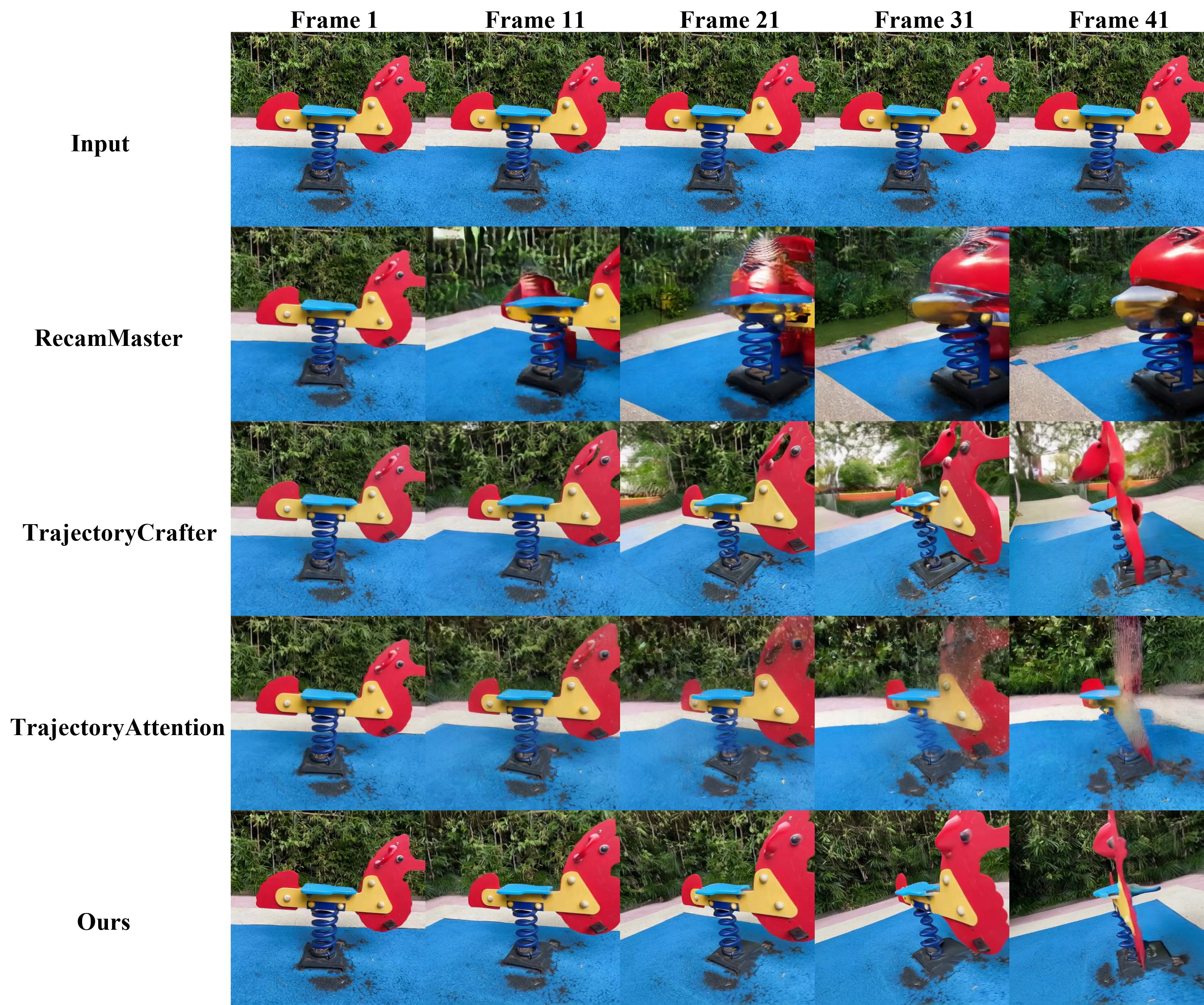}
    \end{subfigure}
    \caption{Comparison of EX-4D with state-of-the-art methods.}
    \label{fig:extra_results5}
\end{figure}

\begin{figure}[ht]
    \centering
    \begin{subfigure}{0.9\linewidth}
        \includegraphics[width=\linewidth]{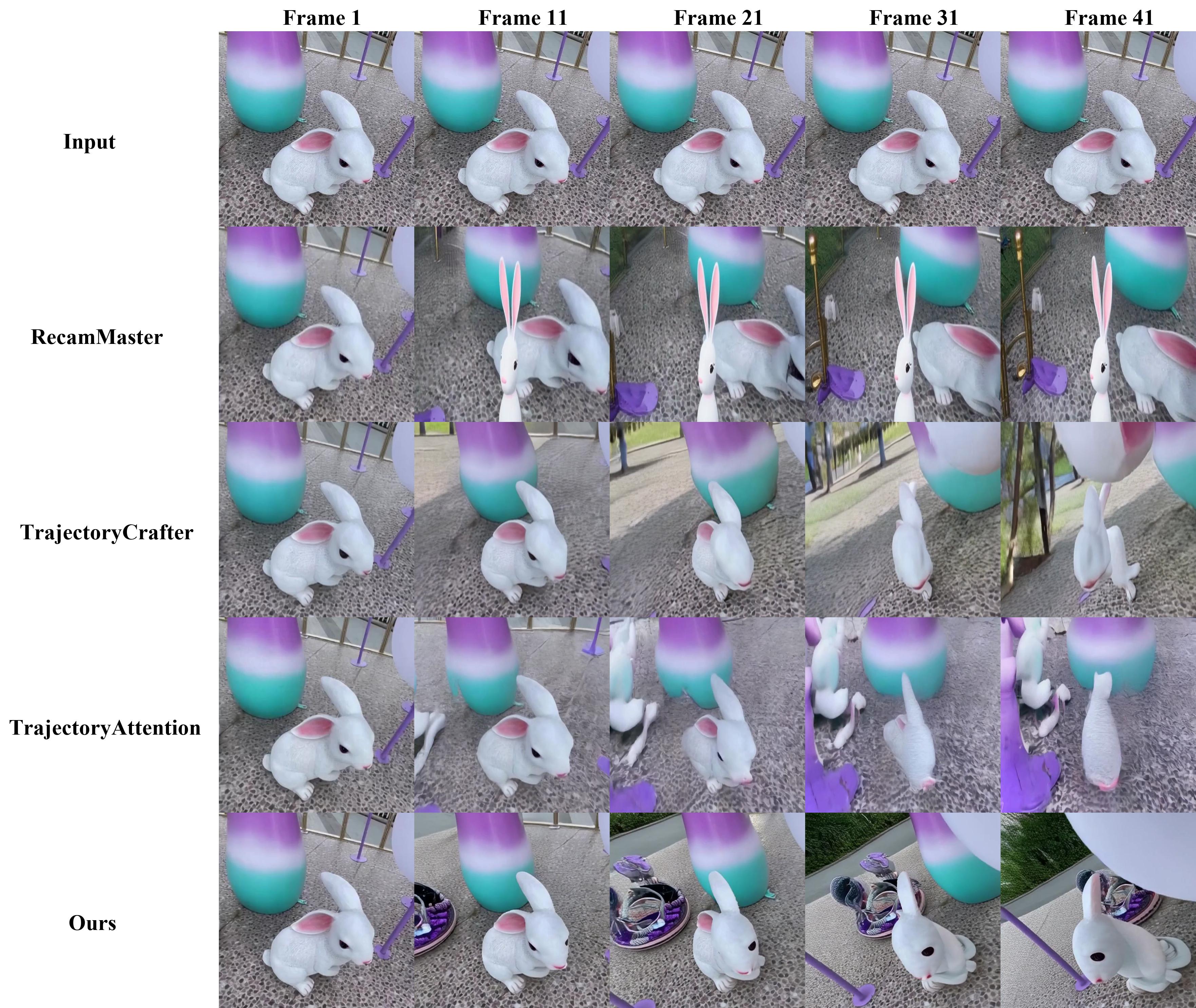}
    \end{subfigure}
    \begin{subfigure}{0.9\linewidth}
        \includegraphics[width=\linewidth]{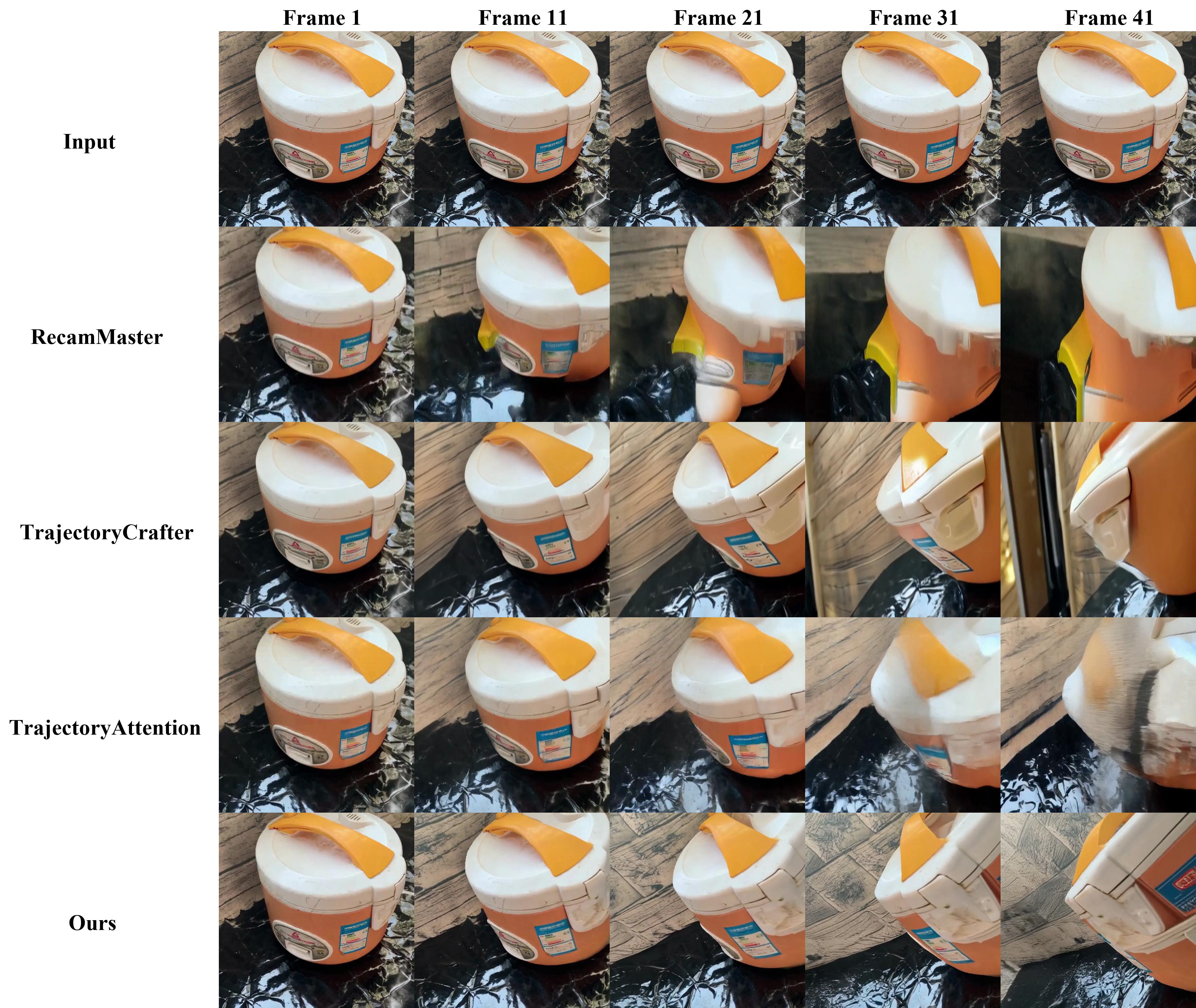}
    \end{subfigure}
    \caption{Comparison of EX-4D with state-of-the-art methods.}
    \label{fig:extra_results6}
\end{figure}

We provide additional visual comparisons between our EX-4D method and state-of-the-art approaches. Fig.~\ref{fig:extra_results1}, Fig.~\ref{fig:extra_results2}. Fig.~\ref{fig:extra_results3}, Fig.~\ref{fig:extra_results4}, Fig.~\ref{fig:extra_results5} and Fig.~\ref{fig:extra_results6} show results across diverse scenes and challenging camera trajectories.

\subsection{Demo Video}
We provide a demo video showcasing the capabilities of our EX-4D framework \phy{in the supplementary file}. \phy{Using scenes synthesized by SOTA video generation models such as Veo3, Sora, and Kling, we generate highly physically consistent novel views under extreme and complex camera trajectories.} The results highlight the effectiveness of our approach in generating high-quality, temporally consistent videos under extreme viewpoints.


\end{document}